\documentclass[10pt,twocolumn,letterpaper]{article}

\usepackage{3dv}
\usepackage{times}
\usepackage{epsfig}
\usepackage{graphicx}
\usepackage{amsmath}
\usepackage{amssymb}
\usepackage{commath}
\usepackage{multicol}
\usepackage{capt-of}
\usepackage{pifont}
\usepackage{stfloats}
\usepackage{enumitem} %
\usepackage{comment} %
\usepackage{balance}  %
\usepackage{float}
\usepackage{booktabs}

\usepackage[pagebackref=true,breaklinks=true,colorlinks,bookmarks=false]{hyperref}
\usepackage[numbers,sort,compress]{natbib}
\usepackage{newfloat}
\DeclareFloatingEnvironment[
    fileext=los,
    listname={List of appendix figures},
    name=Figure,
    placement=tbhp,
    within=section,
]{sfig}

\makeatletter
\def\ps@myheadings{%
    \let\@oddfoot\@empty\let\@evenfoot\@empty
    \def\@evenhead{\thepage\hfil\slshape\leftmark}%
    \def\@oddhead{{\slshape\rightmark}\hfil\thepage}%
    \let\@mkboth\@gobbletwo
    \let\sectionmark\@gobble
    \let\subsectionmark\@gobble
    }
  \if@titlepage
  \renewcommand\maketitle{\begin{titlepage}%
  \let\footnotesize\small
  \let\footnoterule\relax
  \let \footnote \thanks
  \null\vfil
  \vskip 60\p@
  \begin{center}%
    {\LARGE \@title \par}%
    \vskip 3em%
    {\large
     \lineskip .75em%
      \begin{tabular}[t]{c}%
        \@author
      \end{tabular}\par}%
      \vskip 1.5em%
    {\large \@date \par}%
  \end{center}\par
  \@thanks
  \vfil\null
  \end{titlepage}%
  \setcounter{footnote}{0}%
}
\else
\renewcommand\maketitle{\par
  \begingroup
    \renewcommand\thefootnote{\@fnsymbol\c@footnote}%
    \def\@makefnmark{\rlap{\@textsuperscript{\normalfont\@thefnmark}}}%
    \long\def\@makefntext##1{\parindent 1em\noindent
            \hb@xt@1.8em{%
                \hss\@textsuperscript{\normalfont\@thefnmark}}##1}%
    \if@twocolumn
      \ifnum \col@number=\@ne
        \@maketitle
      \else
        \twocolumn[\@maketitle]%
      \fi
    \else
      \newpage
      \global\@topnum\z@   %
      \@maketitle
    \fi
    \thispagestyle{plain}\@thanks
  \endgroup
  \setcounter{footnote}{0}%
}
\makeatother

\threedvfinalcopy %

\def\threedvPaperID{*} %

\ifthreedvfinal\fi

\begin{document}

\title{\TITLE}
\author{Zicong Fan$^{1, 2}$ \quad Adrian Spurr$^{1}$ \quad Muhammed Kocabas$^{1,2}$ \quad Siyu Tang$^{1}$ \\Michael J. Black$^{2}$ \quad Otmar Hilliges$^{1}$\vspace{0.1cm} \\
 $^1$ETH Z{\"u}rich, Switzerland \quad
 $^2$Max Planck Institute for Intelligent Systems, T{\"u}bingen \\
\href{https://zc-alexfan.github.io/digit}{https://zc-alexfan.github.io/digit}
}

\definecolor{mypurple}{RGB}{0, 128, 255}
\newcommand{\myparagraph}[1]{\noindent\textbf{#1.}}

\newcommand{\ra}{{\color{blue}\textbf{R1}}}
\newcommand{\rb}{{\color{green}\textbf{R2}}}
\newcommand{\rc}{{\color{magenta}\textbf{R3}}}

\newcommand{\add}[1]{{\color{blue}#1}}

\newcommand{\reffig}[1]{Fig.~\ref{#1}}
\newcommand{\reftab}[1]{Table~\ref{#1}}
\newcommand{\refeq}[1]{Eq.~\ref{#1}}
\newcommand{\ccite}[1]{~\cite{#1}}

\newcommand{\M}[1]{\mathbf{#1}} %
\newcommand{\V}[1]{\mathbf{#1}} %

\newcommand{\R}[0]{\rm I\!R}
\newcommand{\E}[0]{\rm I\!E}
\newcommand{\loss}[0]{\mathcal{L}}

\newcommand{\fullname}[0]{DIGIT (DIsambiGuating hands in InTeraction)}
\newcommand{\shortname}[0]{DIGIT}
\newcommand{\INTERHAND}[0]{InterHand2.6M \cite{interhand} }
\newcommand{\INTER}[0]{InterHand2.6M \cite{interhand}} %
\newcommand{\suppl}[0]{Sup.~Mat}

\newcommand{\SLASHTXT}[0]{The left and right of the slash are for the single and interacting images.}

\newcommand{\TITLE}[0]{Learning to Disambiguate Strongly Interacting Hands\\ via Probabilistic Per-Pixel Part Segmentation}

\twocolumn[{%
\renewcommand\twocolumn[1][]{#1}%
\maketitle
\begin{center}
  \newcommand{\teaserwidth}{\textwidth}
  \vspace{-0.7cm}
  \centerline{\includegraphics[width=0.95\linewidth]{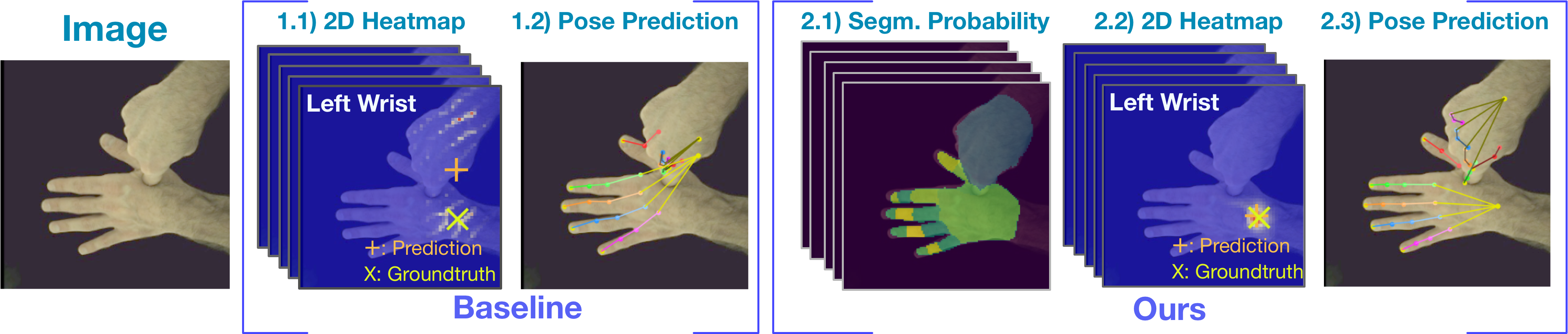}}
    \captionof{figure}{
    When estimating the 3D pose of interacting hands, state-of-the-art methods struggle to disambiguate the appearance of the two hands and their parts. In this example, the baseline fails to differentiate the left and the right wrists (1.1), resulting in erroneous pose estimation (1.2). Our model, DIGIT, reduces the ambiguity by predicting and leveraging a probabilistic part segmentation volume (2.1) to produce reliable pose estimates even when the two hands are in direct contact and under significant occlusion (2.2, 2.3).}
    \label{fig: teaser}
\end{center}%
}]

\thispagestyle{empty}

\begin{abstract}
\vspace{-4mm}
In natural conversation and interaction, our hands often overlap or are in contact with each other. Due to the homogeneous appearance of hands, this makes estimating the 3D pose of interacting hands from images difficult. 
In this paper we demonstrate that self-similarity, and the resulting ambiguities in assigning pixel observations to the respective hands and their parts, is a major cause of the final 3D pose error. 
Motivated by this insight, we propose \shortname, a novel method for estimating the 3D poses of two interacting hands from a single monocular image. 
The method consists of two interwoven branches that process the input imagery into a per-pixel semantic part segmentation mask and a visual feature volume. 
In contrast to prior work, we do not decouple the segmentation from the pose estimation stage, but rather leverage the per-pixel probabilities directly in the downstream pose estimation task. 
To do so, the part probabilities are merged with the visual features and processed via fully-convolutional layers.
We experimentally show that the proposed approach achieves new state-of-the-art performance on the \INTERHAND dataset. 
We provide detailed ablation studies to demonstrate the efficacy of our method and to provide insights into how the modelling of pixel ownership affects 3D hand pose estimation. 
\end{abstract}

\section{Introduction}
Hands are our primary means for manipulating objects and play a key role in communication.
Consequently, a method for estimating 3D hand pose from monocular images would have many applications in human-computer interaction, AR/VR, and robotics. 
We often use both hands in a concerted manner and, as a consequence, our hands are often close to, or in contact with, each other.
Most 3D hand pose estimation methods assume inputs containing a single hand \ccite{iqbal2018hand, DBLP:conf/cvpr/MuellerBSM0CT18, DBLP:conf/cvpr/Spurr0PH18, DBLP:conf/cvpr/TekinBP19, DBLP:conf/cvpr/YangY19, DBLP:conf/iccv/ZimmermannB17, DBLP:conf/cvpr/DoostiNMC20, DBLP:conf/iccv/CaiGLCCYM19, DBLP:conf/eccv/FanLW20, interhand, spurr2020eccv}. This is for good reasons: hands display a large amount of self-similarity and are very dexterous. 
This leads to self-occlusion, which, together with the inherent depth-ambiguities, results in a challenging pose reconstruction problem. Estimating two interacting hands is even more difficult due to the self-similar appearance and complex occlusion patterns, where often large areas of the hands are unobservable. 

Recently, Moon \etal\ccite{interhand} proposed a large-scale annotated dataset, captured via a massive multi-view setup, enabling the study of the 3D interacting hand pose estimation task. Their method shows feasibility but struggles with interacting hands. One of the main challenges in the task is ambiguities caused by the relatively homogeneous appearance of hands and fingers. Even the fingers of a single hand can be difficult to tell apart if only parts of the hand are visible. Considering hands in close interaction only makes this problem more pronounced. 
Consider the example in \reffig{fig: teaser}. Here an interacting hand pose estimator struggles to disambiguate the left and the right wrists, reflected in a bi-modal heatmap, resulting in a poor 3D pose estimate.

To address this problem, we introduce \textit{\fullname}, a simple yet effective method for learning-based reconstruction of 3D hand poses for interacting hands. 
The key insight is to explicitly reason about the per-pixel segmentation of the images into the separate hands and their parts, thus assigning ownership of each pixel to a specific part of one of the hands. 
We show that this reduces the ambiguities brought on by the self-similarity of hands and, in turn, significantly improves the accuracy of 3D pose estimates.
While prior work on hand-\ccite{DBLP:conf/cvpr/BoukhaymaBT19,DBLP:conf/iccv/ZimmermannB17} and body-pose estimation\ccite{omran2018neural, pavlakos2018learning, zanfir2020weakly} and hand-tracking\ccite{mueller2019real, han2020megatrack, wang2020rgb2hands} have leveraged some form of segmentation, most often silhouettes or per-pixel masks, this is typically done as a pre-processing step. 
In contrast, our ablations show that integrating a semantic segmentation branch into an end-to-end trained architecture already increases pose estimation accuracy. We also show that leveraging the per-pixel probabilities, rather than class labels, alongside the image features further improves 3D pose estimation accuracy. 
Finally, our experiments reveal that our method helps to disambiguate interacting hands, and improves the accuracy in estimating the relative positions between hands, via a reduction of uncertainty due to self-similarity.

\begin{figure*}[t]
\centering
  \includegraphics[width=0.8\linewidth]{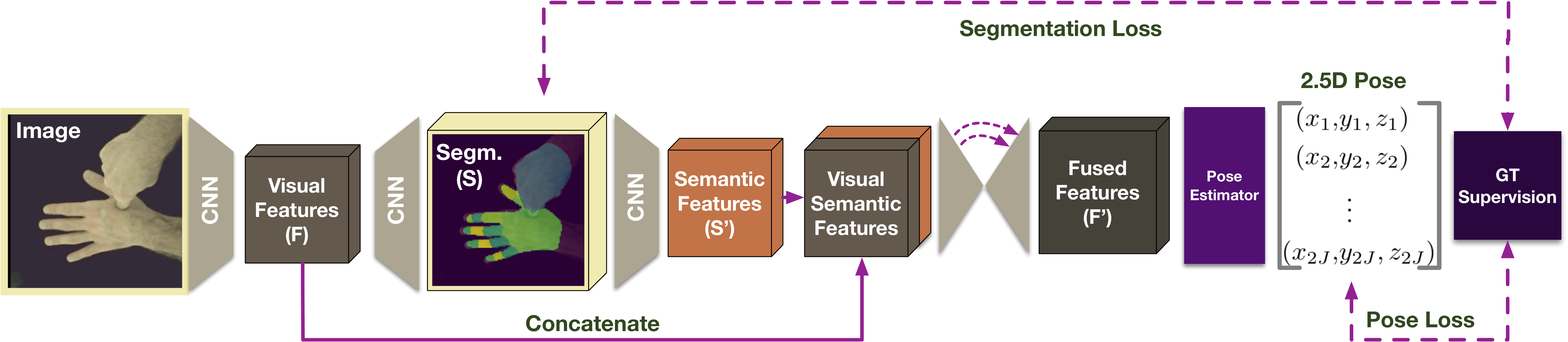}
  \caption{\textbf{An illustration of our hand pose estimation model (\shortname).} Given an image, \shortname\ extracts visual features ($\M{F}$) and predicts a part segmentation probability volume ($\M{S}$). The segmentation volume is projected into latent semantic features ($\M{S'}$). The visual features ($\M{F}$) and the semantic features ($\M{S'}$) are fused across multiple scales and are used for interacting hand pose estimation (illustrated in \reffig{fig: pose_est}).}
  \label{fig: pipeline}
  \vspace{-4mm}
\end{figure*}

More precisely, \shortname\ is an end-to-end trainable network architecture (see \reffig{fig: pipeline}) that uses two separate, but interwoven, branches for the tasks of semantic segmentation and pose estimation respectively. Importantly, the output of the segmentation branch is per-pixel logits (\ie, the full probability distribution) rather than the more commonly used discrete class labels so that the uncertainty in part segmentation prediction is preserved. These probabilities are then merged with the visual features and processed via fully convolutional layers to attain a fused feature representation that is ultimately used for the final pose estimates. The network is supervised via a 3D pose estimation loss and a semantic segmentation loss. We show that all design choices are necessary in order to attain the best performing architecture in ablation studies and that the final proposed method achieves state-of-the-art performance on the \INTERHAND dataset.   
In summary, we contribute:
\begin{enumerate}[noitemsep]
\item An analysis showing that SOTA hand-pose methods are sensitive to self-occlusions and ambiguities brought on by interacting hands
\item A simple yet effective end-to-end trainable architecture for 3D hand pose estimation from monocular images that depict two hands, often under self-contact. 
\item An approach to incorporate a semantic part-segmentation network and means to combine the per-pixel probabilities with visual features for the final task of 3D pose estimation.
\item Detailed ablation studies of our method revealing a reduction of ambiguities due to the self-similarity of interacting hands, and
improvements in estimating 3D hand pose and the relative position between hands.
\item Our method achieves state-of-the-art performance on the \INTERHAND dataset across all metrics.
\end{enumerate}

\section{Related work}
Here we briefly review related work in monocular hand pose estimation, reconstruction of the 3D pose of bi-manual interaction, and the use of segmentation in related tasks.

\myparagraph{Monocular 3D hand pose estimation}
Monocular RGB 3D hand pose estimation has a long history beginning with Rehg and Kanade \cite{Rehg:1994}.
\textit{Surface-based} approaches estimate dense hand surfaces by either fitting a hand model to observations or by regressing model parameters directly from pixels \ccite{heap1996towards, panteleris2018using, lu2003using, de2011model, DBLP:journals/cviu/IqbalDYK0G18, DBLP:conf/iccv/OikonomidisKA11, mano, DBLP:conf/cvpr/HassonTBLPS20, moon2020deephandmesh, DBLP:conf/iccv/ZhangLMZZ19, Moon_2020_ECCV_I2L-MeshNet, DBLP:conf/cvpr/HassonVTKBLS19, Rehg:1994, DBLP:conf/cvpr/KulonGKBZ20, ge20193d}. 
More closely related to ours are \textit{keypoint-based} approaches, that regress the 3D joint positions \ccite{iqbal2018hand, DBLP:conf/cvpr/MuellerBSM0CT18, DBLP:conf/cvpr/Spurr0PH18, DBLP:conf/cvpr/TekinBP19, DBLP:conf/cvpr/YangY19, DBLP:conf/iccv/ZimmermannB17, DBLP:conf/cvpr/DoostiNMC20, DBLP:conf/iccv/CaiGLCCYM19, DBLP:conf/eccv/FanLW20, interhand, spurr2020eccv}. 
For example, Zimmermann \etal\ccite{DBLP:conf/iccv/ZimmermannB17} propose the first convolutional network for RGB hand pose estimation. Iqbal \etal\ccite{iqbal2018hand} introduce a 2.5D representation, allowing training on in-the-wild 2D annotation. 
However, all of the above approaches assume single-hand images.  
Recently, Moon \etal\ccite{interhand} introduce a large-scale dataset and a 3D hand pose estimator for images with interacting hands. 
In our work, we show that existing approaches struggle with occlusions and appearance ambiguities. To this end, we propose a simple yet effective method that can better disambiguate strongly interacting hands and thus improves interacting hand pose estimation.

\myparagraph{Interacting hand tracking and pose estimation}
\label{sec: related_work_hand_track_pose}
Model-based approaches to tracking of interacting hands have been proposed\ccite{oikonomidis2012tracking, ballan2012motion, tzionas2016capturing, mueller2019real, wang2020rgb2hands, smith2020constraining} as well as \emph{multi-view} methods to reconstruct the pose of interacting hands\ccite{bootstrapping, interhand, he2020epipolar}.
Oikonomidis \etal\ccite{oikonomidis2012tracking} provide a formulation to track interacting hands using Particle Swarm Optimization from RGBD videos. Ballan \etal\ccite{ballan2012motion} introduce an offline method to capture hand motion during hand-hand and hand-object interaction in a multi-camera setup. Tzionas \etal\ccite{tzionas2016capturing} extend the idea in\ccite{ballan2012motion} with a physical model. Mueller \etal\ccite{mueller2019real} and Wang \etal\ccite{wang2020rgb2hands} propose interacting hand tracking methods by predicting left/right-hand silhouettes and correspondence masks used in a post-processing energy minimization step. Smith \etal \ccite{smith2020constraining} propose a multi-view system that constrains a vision-based tracking algorithm with a physical model.
In 3D interacting hand pose estimation, Simon \etal\ccite{bootstrapping} propose a multi-view bootstrapping technique to triangulate full-body 2D keypoints into 3D. He \etal\ccite{he2020epipolar} incorporate epipolar geometry into a transformer network\ccite{vaswani2017attention}.
Moon \etal\ccite{interhand} propose a large-scale dataset and a model for interacting hand pose estimation.
Moon \etal\ccite{interhand} is the most closely related to this paper as it estimates 3D hand poses of interacting hands from a monocular RGB image. 
Compared to hand tracking\ccite{oikonomidis2012tracking, ballan2012motion, tzionas2016capturing, mueller2019real, wang2020rgb2hands, smith2020constraining}, we do not require image sequences. In contrast to the existing interacting pose estimation methods\ccite{interhand, bootstrapping, he2020epipolar, lin2021two}, we explicitly model uncertainty caused by appearance ambiguity in interacting hands and we do not require a multi-view supervision\ccite{bootstrapping, he2020epipolar}.

\myparagraph{Segmentation in pose estimation}
Segmentation has been used in 3D hand pose estimation, 3D human pose estimation and hand tracking and can be grouped into four categories: as a localization step\ccite{zhang2020weakly, DBLP:conf/iccv/ZimmermannB17, oberweger2015hands, oberweger2017deepprior++, athitsos2003estimating, kang2017hand}, as a training loss\ccite{DBLP:conf/cvpr/BoukhaymaBT19, DBLP:conf/cvpr/BaekKK19}, as an optimization term\ccite{che2018dynamic, mueller2019real, wang2020rgb2hands}, or as an intermediate representation\ccite{omran2018neural, zanfir2020weakly, pavlakos2018learning, neverova2017hand, chen2018shpr}. Most single hand pose estimation approaches follow Zimmermann \etal\ccite{DBLP:conf/iccv/ZimmermannB17} in localizing a hand in an image by predicting the hand silhouette, which is used to crop the input image before performing pose estimation. Boukhayma \etal\ccite{DBLP:conf/cvpr/BoukhaymaBT19} predict a dense hand surface and use a neural rendering technique to obtain a silhouette loss. 
In contrast, we leverage part segmentation to explicitly address self-similarity in interacting hand pose estimation. 
In depth-based hand pose estimation, \ccite{neverova2017hand, chen2018shpr} apply segmentation on 3D point clouds for single-hand pose estimation. Our focus is on interacting hands from RGB images, which is challenging because fingers of the hands share very similar texture and there is depth ambiguity when estimating from RGB images. 
In tracking interacting hands, left and right-hand masks can be predicted from either depth images\ccite{mueller2019real} or monocular RGB images\ccite{wang2020rgb2hands}, which are used in an optimization-based post-processing step. 
Our method neither assumes RGB nor depth image sequences. 
In 3D human pose and shape estimation, existing methods predict part segmentation maps\ccite{omran2018neural, zanfir2020weakly} or silhouettes\ccite{pavlakos2018learning} from RGB images and use the predicted masks as an intermediate representation. Specifically, they decouple the image-to-pose problem into image-to-segmentation and segmentation-to-pose. The two models of the subtasks are trained separately. In contrast, our method trains image-to-pose in an end-to-end fashion. The most closely related method is that of Omran \etal\ccite{omran2018neural}. 
In addition to not training the networks end-to-end, they use discrete part segments while we preserve uncertainty with a probabilistic segmentation map. We show that end-to-end training and the use of probabilistic maps significantly reduce hand pose estimation errors.

\section{Method}

\subsection{Overview}
At the core of \shortname\ lies the observation that the self-similarity between joints in two hands, which is especially pronounced during hand-to-hand interaction, is a major source of error in monocular 3D hand pose estimation with interacting hands.
Our experiments (see \reffig{fig: iou_mpjpe}) show that standard approaches do not have a mechanism to cope with such ambiguity. 
Embracing this challenge, we propose a simple yet effective framework for interacting hand pose estimation by modelling per-pixel ownership via probabilistic part segmentation maps. Our method leverages both visual features and distinctive semantic features to address the ambiguity caused by self-similarity.

In contrast to prior work, which separates the image-to-segmentation and segmentation-to-pose steps \ccite{omran2018neural, zanfir2020weakly, pavlakos2018learning}, we propose a holistic approach that is trained to jointly reason about the pixel-to-part assignment and 3D joint locations. 
In particular, given an input image, our model identifies individual parts of each hand in the form of probabilistic segmentation maps that are used to encourage the local influence of visual features for estimating the corresponding 3D joint.
We fuse the probabilistic segmentation maps and the visual feature maps across multiple scales using a convolutional fusion layer. 
Our experiments show that end-to-end training and the use of probabilistic segmentation maps significantly improve hand pose estimation.

\subsection{Segmentation-aware pose estimation}

\begin{figure}[t]
\centering
  \includegraphics[width=0.8\linewidth]{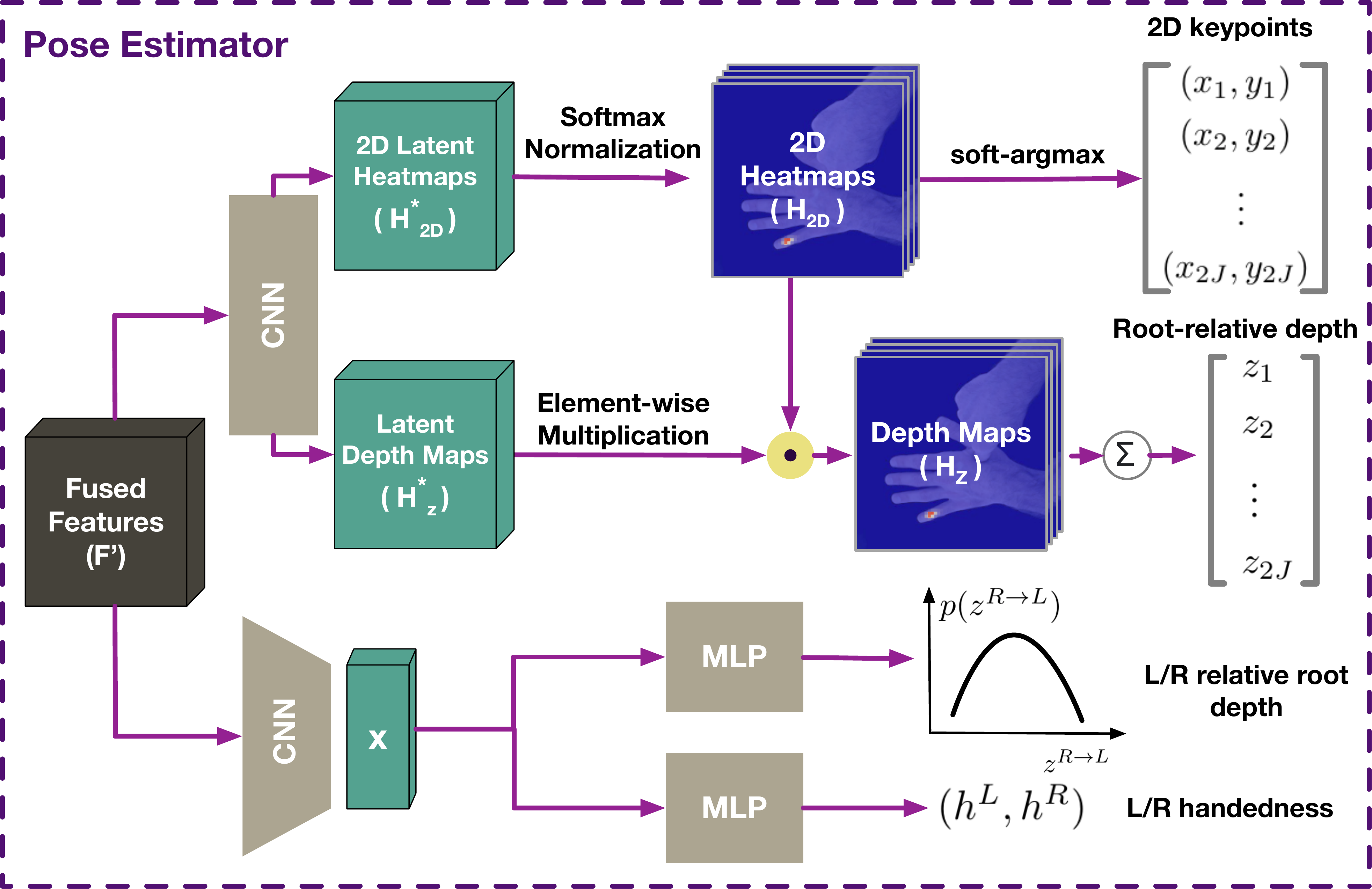}
  \caption{\textbf{Our interacting hand pose estimator.}}
  \label{fig: pose_est}
\end{figure}
\begin{figure}
\centering
  \includegraphics[width=0.23\linewidth]{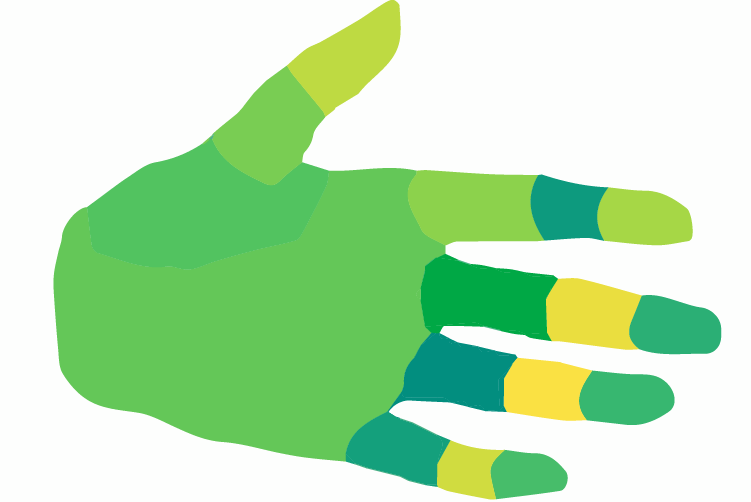}
  \caption{\textbf{Part segmentation classes of our model.} Each hand is partitioned into 16 parts as shown above in different colors. Including the background and parts from both hands, there are 33 classes for part segmentation. Here we show one hand for brevity.}
  \label{fig: part_label}
  \vspace{-4mm}
\end{figure}

Figure \ref{fig: pipeline} illustrates the main components of our framework.
Given an image region $\M{I}\in \R^{W_I\times H_I\times 3}$, cropped by a bounding box including all hands, the goal of our model is to estimate the 3D hand pose $\M{P}_{3D}\in \R^{2J\times 3}$ in the camera coordinate for $2J$ joints where $J$ is the number of joints in one hand.
In particular, we first extract a feature map $\M{F}\in \R^{W_F\times H_F\times D_F}$ from the image $\M{I}$ using a CNN backbone network to provide visual features for pose estimation and part segmentation. Here $W_F$, $H_F$, and $D_F$ denote the width, height and channel dimensions of the feature map.

\myparagraph{Probabilistic segmentation} Since there is an inherent self-similarity between different parts of the hands, we learn a part segmentation network to predict a probabilistic segmentation volume $\M{S}\in \R^{W_S\times H_S\times C}$, which is directly supervised by groundtruth part segmentation maps. 
Each pixel on the segmentation volume $\M{S}$ is a channel of probability logits over $C$ classes where $C$ is the number of categories including the parts of the two hands and the background (see \reffig{fig: part_label}).
Note that to preserve the uncertainty in segmentation prediction, we do not pick the class with the highest response among the $C$ classes for each pixel in the segmentation volume. For display purposes only, we show the class with the highest probability in \reffig{fig: pipeline}.
Finally, since the segmentation volume $\M{S}$ has a higher resolution than $\M{F}$, we perform a series of convolution and downsampling operations to obtain semantic features $\M{S'}\in \R^{W_F\times H_F\times D_S}$ where $D_S$ is the channel dimension.

\myparagraph{Visual semantic fusion}
The visual features $\M{F}$ and the semantic features $\M{S'}$ are concatenated along the channel dimension to provide rich visual cues for estimating accurate 3D hand poses and distinctive semantic features for avoiding appearance ambiguity. 
However, a naive concatenation does not provide global context from the semantic features. 
Therefore, we fuse the visual and semantic features across different scales using a custom and lightweight UNet\ccite{ronneberger2015u} to obtain a fused feature map $\M{F'} \in \R^{W_F\times H_F\times (D_F + D_S)}$ for pose estimation (see \suppl. for the UNet details).

\myparagraph{Interacting hand pose estimation}
To estimate the final 3D hand pose of both hands, we learn a function $\mathcal{F}: \M{F'} \rightarrow \M{P}_{2.5D}$ that maps the fused feature map $\M{F'}$ to 2.5D pose. 
The 2.5D representation $\M{P}_{2.5D}\in \R^{2J\times 3}$ consists of individual 2.5D joints $(x_i, y_i, z_i)\in \R^3$ where $(x_i, y_i)$ is the 2D projection of the 3D joint $(X_i, Y_i, Z_i)\in \R^3$, and $z_i=Z_i - Z_{root(i)}$. 
The notation $root(i)$ denotes the hand root of joint $i$.
During inference, 3D poses can be recovered by applying inverse perspective projection on $(x_i, y_i)$ using the depth estimation $Z_i$.
To model the function $\mathcal{F}$, we use a custom estimator inspired by\ccite{iqbal2018hand, interhand}. We found that the 2.5D representation of\ccite{iqbal2018hand} performs equally well to the interacting pose estimator by Moon \etal\ccite{interhand}, while being more memory-efficient due to the lack of a requirement for a volumetric heatmap representation (see \suppl.). 

Figure \ref{fig: pose_est} shows a schematic of our proposed pose estimator.
Similar to\ccite{interhand}, our model estimates the handedness $(h^L, h^R)\in [0, 1]^2$, the 2.5D left and right-hand pose $\M{P}_{2.5D}$, and the right hand-relative left-hand depth $z^{R\rightarrow L}\in \R$, where $L$ and $R$ denote left and right hands. 
Since our model predicts 2.5D joints $(x_i, y_i, z_i)$, and converting from 2.5D to 3D requires an inverse perspective projection, we need to estimate the depth $Z_i$ for a joint $i$ by $Z_i = z_i+ Z_{root(i)}$. The root relative depth $z^{R\rightarrow L}$ is used to obtain the left-hand root depth when both hands are present.

\myparagraph{Handedness and relative root depth} The handedness  $(h^L, h^R)$ detects the presence of the two hands and $z^{R\rightarrow L}$ measures the depth of the left root relative to the right root.
We repeatedly convolve and downsample $\M{F'}$ to a latent vector $\V{x}$, which is used to estimate $(h^L, h^R)$ and $z^{R\rightarrow L}$ by two separate multi-layer perception (MLP) networks.
For $z^{R\rightarrow L}$, we use the MLP to estimate a 1D heatmap $\V{p}\in \R^{D_z}$ that is softmax-normalized, representing the probability distribution over $D_z$ possible values for $z^{R\rightarrow L}$. The final relative depth $z^{R\rightarrow L}$ is obtained by
\begin{align}\label{eq: z_relative}
    z^{R\rightarrow L} = \sum_{k=0}^{D_z-1} k\:\V{p}[k].
\end{align}

\myparagraph{2.5D hand pose estimator ($\mathcal{F}$)} Inspired by\ccite{iqbal2018hand}, our pose estimator predicts the latent 2D heatmap $\M{H}^*_{2D} \in \R^{W_F\times H_F \times 2J}$ for the 2D joint locations and the latent root-relative depth map $\M{H}^*_{z} \in \R^{W_F\times H_F \times 2J}$ for the root-relative depth of each joint. 
The heatmap $\M{H}^*_{2D}$ is spatially softmax-normalized to a probability map $\M{H}_{2D} \in \R^{W_F\times H_F \times 2J}$. Since $\M{H}_{2D}$ indicates potential 2D joint locations, to focus the depth values on the joint locations, $\M{H}_{2D}$ is element-wise multiplied with the latent depth map $\M{H}^*_{z}$ to obtain the depth map $\M{H}_{z} = \M{H}^*_{z} \odot \M{H}_{2D}$. 
To allow our network to be fully-differentiable, we use soft-argmax\ccite{luvizon2019human} to convert the 2D heatmap $\M{H}_{2D}$ to 2D keypoints $\{(x_i, y_i)\}_{i=0}^{2J}$.
Finally, we sum across the values on each slice of the depth map $\M{H}_{z}$ to obtain the root-relative depth $z_i$ for a joint $i$.

\myparagraph{From 2.5D pose to 3D} 
To convert $\M{P}_{2.5D}$ to 3D pose $\M{P}_{3D}$, following\ccite{interhand}, we apply an inverse perspective projection to map 2D keypoints to 3D camera coordinates by 
\begin{align}
\mathbf{P}_{3 \mathrm{D}}^{\mathrm{L}}&=\Pi\left(\mathbf{T}^{-1} \mathbf{P}_{2.5 \mathrm{D}}^{\mathrm{L}}+\mathbf{Z}^{\mathrm{L}}\right)\\ \mathbf{P}_{3 \mathrm{D}}^{\mathrm{R}}&=\Pi\left(\mathbf{T}^{-1} \mathbf{P}_{2.5 \mathrm{D}}^{\mathrm{R}}+\mathbf{Z}^{\mathrm{R}}\right)
\end{align}

\noindent where $\Pi$ and $\mathbf{T}^{-1}$ are the camera back-projection operation and the inverse affine transformation (undoing cropping and resizing). The projection requires the absolute depth of the left and right roots $\mathbf{Z}^{\mathrm{L}}$ and $\mathbf{Z}^{\mathrm{R}}$ (written in vector form):
\begin{align}
\mathbf{Z}^{\mathrm{L}}&=\left\{\begin{array}{ll}
{[0, 0, z^{\mathrm{L}}]^\intercal,} & \text { if } h^{\mathrm{R}}<0.5 \\
{[0, 0, z^{\mathrm{R}}+z^{\mathrm{R} \rightarrow \mathrm{L}}]^\intercal,} & \text { otherwise, }
\end{array}\right.\\\text{and}\;
\mathbf{Z}^{\mathrm{R}}&=[0, 0 ,z^{\mathrm{R}}]^\intercal
\end{align}
where $z^L$ and $z^R$ are the absolute depth of the roots for the left and the right hands. Following\ccite{interhand}, we use the estimates from RootNet\ccite{moon2019camera} for $z^L$ and $z^R$.

\myparagraph{Training loss}
The loss used to train our model is:
\begin{equation}
    \mathcal{L} = \mathcal{L}_{h} + \mathcal{L}_{2.5D} + \mathcal{L}_{z} + \lambda_s \mathcal{L}_{s}  +  \lambda_b  \mathcal{L}_{b} ,
    \label{eq: total_loss}
\end{equation}
where the terms are the handedness loss $\mathcal{L}_{h}$, the 2.5D hand pose loss $\mathcal{L}_{2.5D}$, the right-hand relative left hand depth loss $\mathcal{L}_{z}$, the segmentation loss $\mathcal{L}_{s}$, and a bone regularization loss $\mathcal{L}_{b}$. 
In particular, we use the multi-label binary cross-entropy loss to supervise the handedness prediction. For segmentation, we use the multi-class cross-entropy loss:
\begin{align}
    \mathcal{L}_s = - \sum_{m=1}^{W_F} \sum_{n=1}^{H_F} \sum_{j=1}^{C} \M{T}_j[m, n] \log (\sigma(\M{S}_j[m, n]))
\end{align}
where $\M{T}_j[m, n]\in \R^C$ is a one-hot vector with 1 positive class and $C-1$ negative classes according to the groundtruth for the segmentation pixel at $(m, n)$ and $\sigma(\cdot)$ is a softmax normalization operation so that $\sum_{j=1}^C\M{S}_j[m, n]=1$.  
We use the L1 loss to supervise the 2.5D hand pose and the relative root depth.

\myparagraph{Kinematic consistency}
In our experiments, we observe that both the method by Moon \etal\ccite{interhand} and our baseline yield asymmetric predictions in terms of bone length between the left and the right hand, due to the appearance ambiguities (with an average difference of 10mm and 8mm in the validation set on the two models respectively).
To encourage more physically plausible predictions we propose a bone vector loss, $\mathcal{L}_{b}=$
\begin{equation}
    \sum_{(i, j)\in \mathcal{E}}\norm{(\M{P}^i_{2.5D} - \M{P}^j_{2.5D}) - (\M{\bar{P}}^i_{2.5D} - \M{\bar{P}}^j_{2.5D})},
\end{equation}
where $(i, j)\in \mathcal{E}$ denotes a bone from the  edge set $\mathcal{E}$ of the directed kinematic tree connecting the groundtruth joints $\M{\bar{P}}^i_{2.5D}$ and $\M{\bar{P}}^j_{2.5D}$. This loss encourages the predicted bones to have a similar length to those in the groundtruth.

\section{Experiments}

Our key observation is that self-similarity, and the resulting ambiguities between joints of the two hands is a major error source for interacting 3D hand pose estimation.
We hypothesize that the ambiguity problem can be alleviated by modelling pixel ownership via part segmentation.
To test the efficacy of our approach, we first compare our model with the state-of-the-art. We investigate the benefits of modelling part segmentation in an ablation study that sheds further light on when and why our approach works.

\myparagraph{Dataset} We evaluate our model using \INTER, which is the only published large-scale dataset for modeling hand-to-hand interaction, including images of both single-hands (SH) and interacting-hands (IH).
We use the 5 frames-per-second (FPS) dataset for our experiments. All models are trained with both single and interacting hand sequences.
We use the final release of\ccite{interhand} for comparison with the state-of-the-art methods in the main paper and include the results of the initial release of\ccite{interhand} in \suppl.
For the ablation experiments, we provide results on the initial release of\ccite{interhand}.
Since \INTERHAND images have a characteristic background, to show generalization, we use the dataset from Tzionas \etal \ccite{tzionas2016capturing} for testing purposes. We use its two-hand subset for the evaluation.

\myparagraph{Evaluation metrics} We use the three metrics from \ccite{interhand} for evaluation. The average precision of handedness estimation (AP) measures handedness prediction accuracy. The root-relative mean per joint position error (MPJPE) measures the error in root-relative 3D hand pose estimation; this is the Euclidean distance between the predicted 3D joint locations and the groundtruth after root alignment. For interacting sequences, the alignment is done on the two hands separately. To measure the performance in estimating the relative position between the left and the right root in interacting sequences, we use the mean relative-root position error (MRRPE).
This is defined as the Euclidean distance between the predicted and the
groundtruth left-hand root position after aligning the left-hand root by the root of the right hand.

\subsection{Implementation details}
We implement our models in PyTorch\ccite{paszke2019pytorch} using an HRNet-W32\ccite{sun2019deep} backbone pre-trained on the ImageNet dataset\ccite{deng2009imagenet}. Following\ccite{interhand}, we crop the hand region using the groundtruth bounding box for both training and testing images and resize the cropped image to $256\times 256$ before feeding it to the network. 
The spatial dimension of the 2D heatmap $\M{H}_{2D}$ and the 2D depth map $\M{H}_{z}$ are $64\times 64$. We obtain groundtruth part segmentation for training our segmentation network by rendering groundtruth hand meshes from \INTERHAND with a neural renderer\ccite{kato2018renderer, kolotouros2018pytorch} using a custom texture map (see Fig.~\ref{fig: part_label}). The details of our network are in \suppl. To balance the loss in \refeq{eq: total_loss}, we choose $\lambda_b = 1.0$ and $\lambda_s=10.0$ based on the average MPJPE for interacting images on the validation set. 
We do not apply $\mathcal{L}_s$ for models without a segmentation network.

\myparagraph{Training procedure}
We train all models with single and interacting hand sequences using the Adam optimizer\ccite{kingma2014adam} with an initial learning rate of $10.0^{-4}$ and a batch size of 64. For experiments comparing to the state-of-the-art, we train our models for 50 epochs and decay the learning rate at epoch 40.
For ablation, unless specified, we train the models for 30 epochs for time efficiency and decay the learning rate at epoch 10 and 20 by a factor of 10.

\subsection{Comparison with the state-of-the-art}

\begin{figure*}[t]
\centering
  \includegraphics[width=0.8\linewidth]{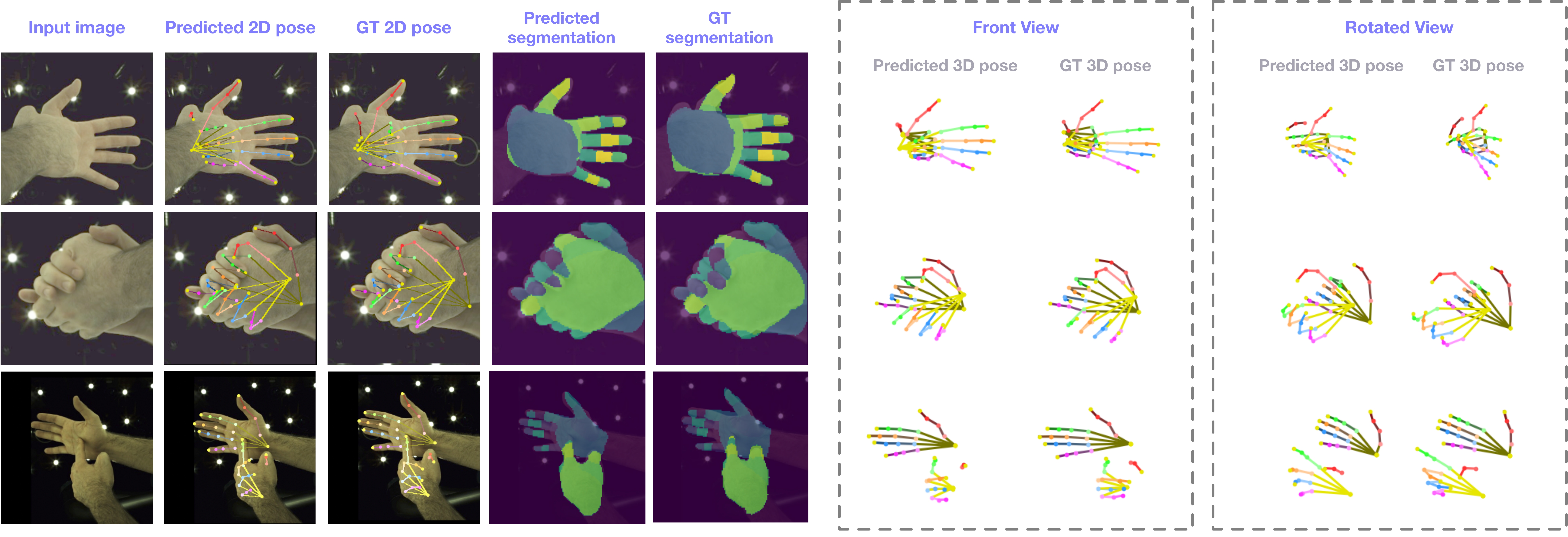}
  \caption{\textbf{Qualitative results from our model.} The lighting is adjusted for display purposes (not model input). Best viewed zoomed in.}
  \label{fig: qualitative}
  \vspace{-4mm}
\end{figure*}

Table \ref{tab: sota_ih} shows the root-relative mean pose per joint error (MPJPE) in millimeters for the interacting hand sequences. Our proposed model outperforms \INTERHAND by 1.86mm on the validation set.
T-tests on the MPJPE values reveal that all differences between ours and the baseline for interacting hands are statistically significant (all $p<10^{-4}$) for both val/test sets.  For interacting hands, it is important to quantify the performance of relative root prediction between the left and right-hand roots, which is reflected by the MRRPE metric. In this metric, our model outperforms \INTERHAND by 4.11mm/3.35mm on the val/test sets.
The average precision for predicting the handedness of \INTER, our baseline, and our proposed model on the entire test set are 99.09, 99.02, and 99.16 percent. 
Figure \ref{fig: qualitative} shows our qualitative results; it shows the input image, 2D keypoint, and the segmentation masks overlaid on the image. The predicted 3D pose is shown in two views.
Although our paper focuses on addressing the ambiguity during hand-to-hand interaction, we also provide the single-hand evaluation in \suppl.\ as a reference.

\begin{figure}[]
\centering
  \includegraphics[width=0.8\linewidth]{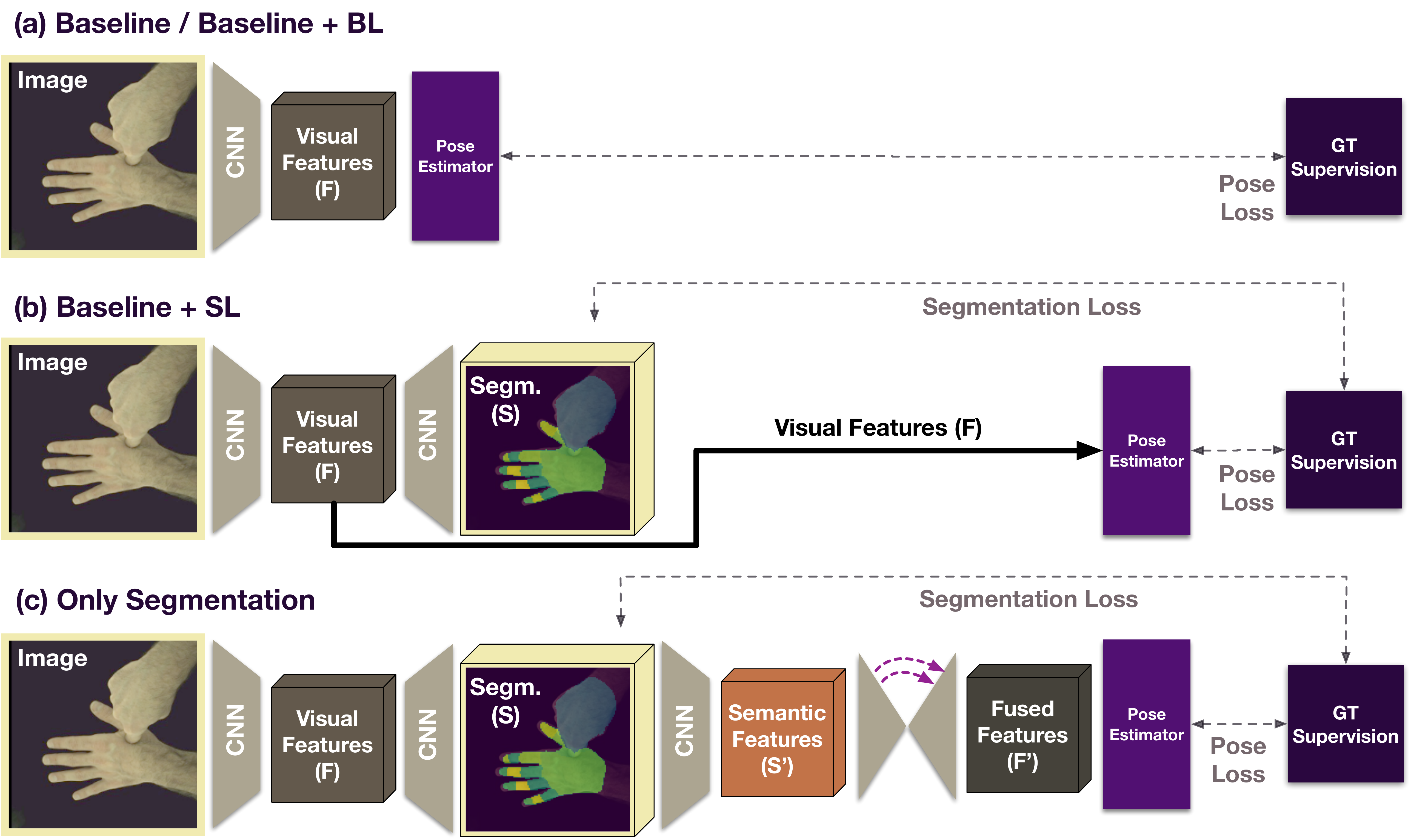}
  \caption{\textbf{Pose estimators for the experiments:} (a) without segmentation, (b) with segmentation but only for applying a loss, (c)  segmentation for pose estimation (no visual features).}
  \label{fig: baseline_sl}
  \vspace{-4mm}
\end{figure}

\begin{table}
\centering
\resizebox{\columnwidth}{!}{%
\begin{tabular}{lcccc}
\toprule
Methods & MPJPE Val  & MRRPE Val  & MPJPE Test  & MRRPE Test \\\hline
\INTERHAND      & 18.58 & 35.64 & 16.02                           & 32.57 \\
Baseline & 17.79 & 33.90 & 15.06 & 31.36                           \\
Ours & \textbf{16.72} & \textbf{31.53} & \textbf{14.27} & \textbf{29.22}\\\hline
\% in improvement over\ccite{interhand} & 10.01 & 11.53 & 10.92 & 10.29\\
\bottomrule
\end{tabular}
}
\caption{\textbf{Comparison with the state-of-the-art.} The MPJPE metric measures the accuracy of interacting 3D hand pose estimation while the MRRPE metric shows the performance of the relative root position between the two hands.}
\label{tab: sota_ih}
\vspace{-4mm}
\end{table}

\subsection{Analysis of our proposed model}
We first provide a qualitative analysis to illustrate how segmentation helps with appearance ambiguity. We then investigate hand pose performance between \INTER, our baseline, and our final model under different degrees of interaction to evaluate how modeling pixel ownership via segmentation helps to reduce errors in hand pose estimation.  Finally, we provide evidence of generalization and detailed analysis on our probabilistic part segmentation.

\begin{figure*}
\centering
  \includegraphics[width=0.8\linewidth]{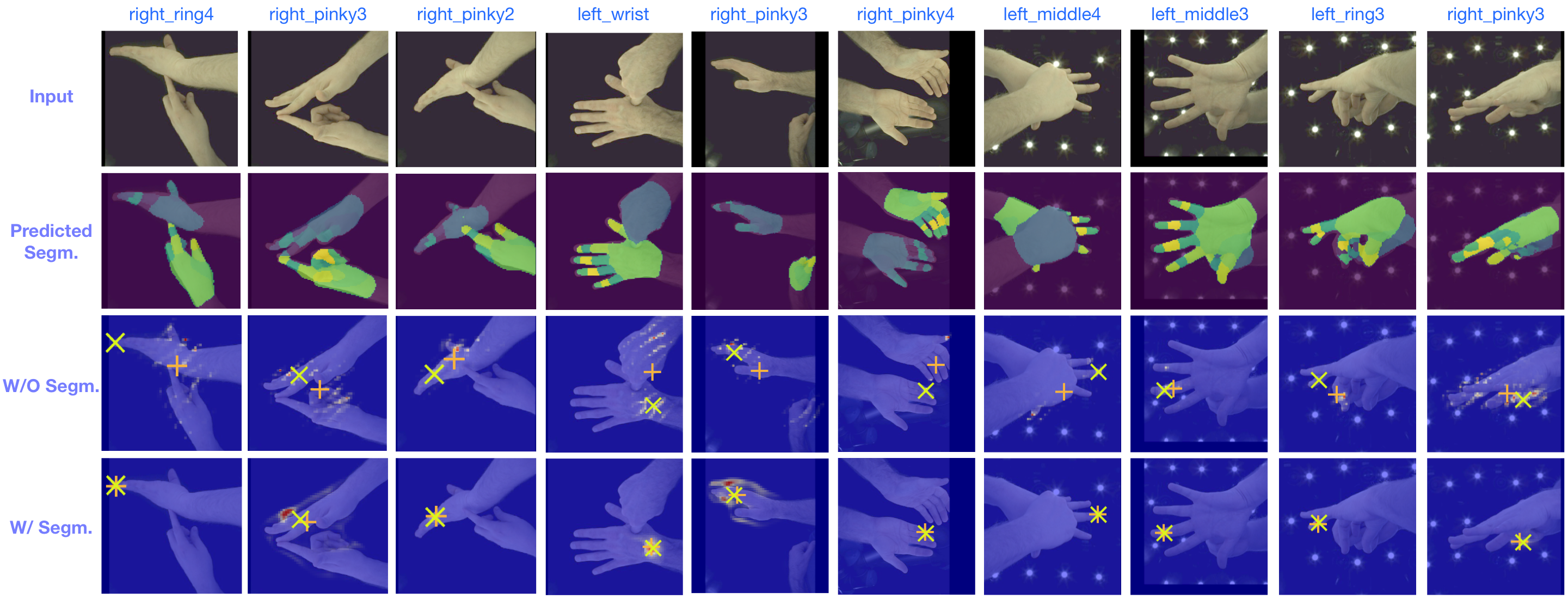}
  \caption{\textbf{Qualitative results of how segmentation reduces appearance ambiguity.} The image lighting is adjusted for display purposes (not model input). The dispersed 2D heatmap issue also arises in the model proposed in \INTERHAND. See \suppl.\ for a more in-depth analysis.
  The skeleton notation is also provided in \suppl. Plus: prediction; Cross: groundtruth. Best viewed zoomed in.}
  \label{fig: qualitative_segm}
  \vspace{-4mm}
\end{figure*}

\begin{figure}[]
\centering
  \includegraphics[width=0.8\linewidth]{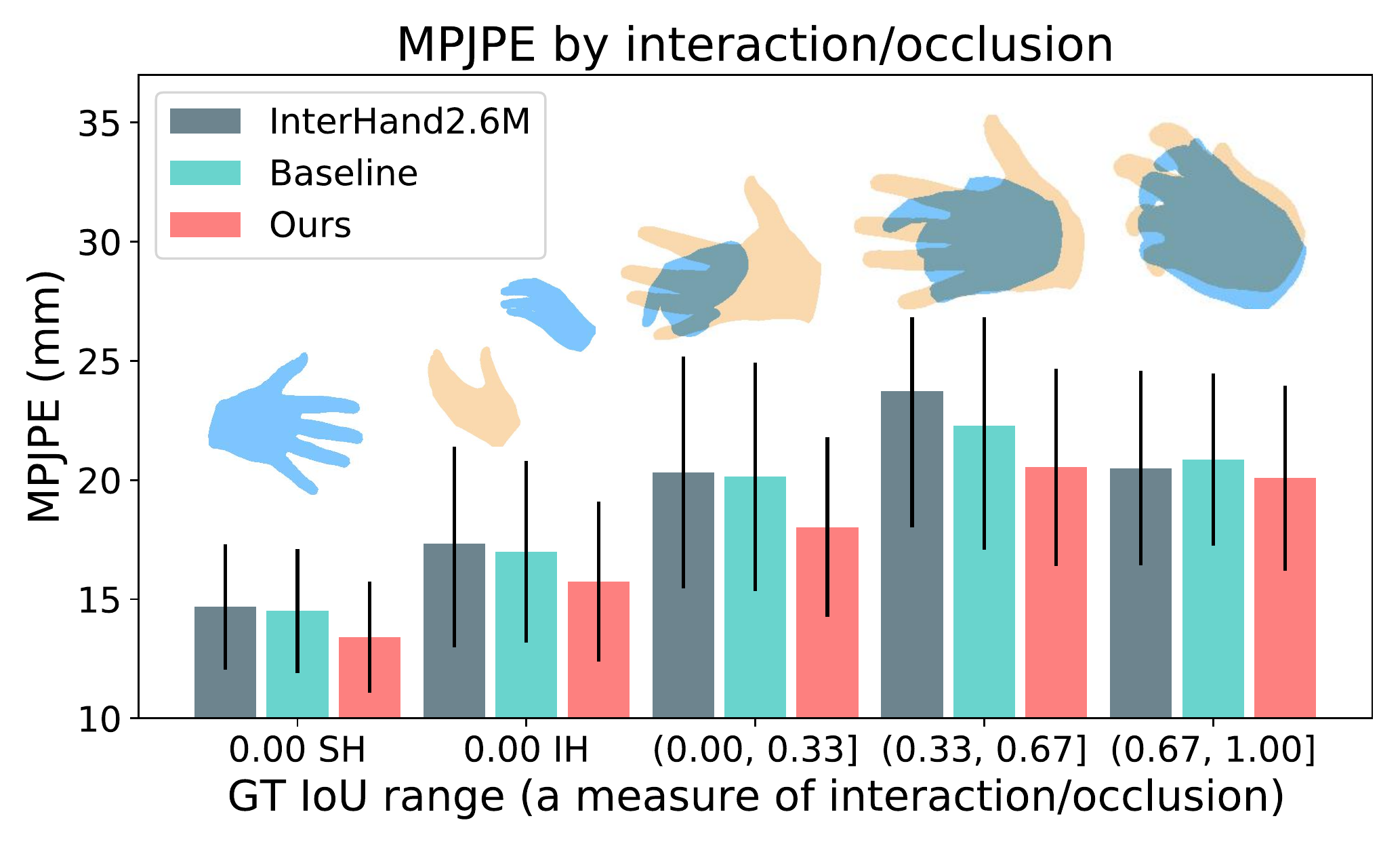}
  \caption{\textbf{Comparing pose estimation performance by the degree of interaction/occlusion.} The IoU between groundtruth left/right masks measures the degree of interaction. SH and IH denote single and interacting images. The left (yellow) and right (blue) hand masks provide interaction examples in each IoU range.}
  \label{fig: iou_mpjpe}
  \vspace{-4mm}
\end{figure}

\myparagraph{Qualitative Analysis}
We investigate how segmentation helps to reduce appearance ambiguity in interacting hand pose estimation.
To build intuition, we provide the 2D estimations of individual joints in \reffig{fig: qualitative_segm}. The cross sign indicates the groundtruth 2D location of the joint of interest and the plus sign is the predicted location. The example in the first column shows that, without segmentation, the baseline model's predictions
contain significant uncertainty due to the presence of the other hand in the image, as indicated by the dispersed 2D heatmap with modes on both hands (see the same behaviour in the \INTERHAND model in \suppl.). 
As a result, the 2D prediction is centered between the hands after the soft-argmax\ccite{luvizon2019human}. 
In contrast, with segmentation, our network 
disambiguates different hands and provides a single-mode estimate. 

\myparagraph{Impact of interaction and occlusion}
We study how pose estimation performance is affected by the degree of interaction.
In particular, we use the IoU between the groundtruth left/right masks (not part segm.) to measure the degree of interaction and occlusion. The higher IoU implies more occlusion.
Figure \ref{fig: iou_mpjpe} compares the results on the entire validation set. The bars show the MPJPE over annotated joints for each IoU range while the half-length of the error bars correspond to 0.5 times (for better display) of MPJPE standard deviation. 
Typical hand masks are shown above the bars of each IoU range.
We also include the errors of single-hand sequences, and observe more errors in interacting hand cases even when the two hands do not intersect, which indicates that the ambiguity applies as long as two hands are present. 
For non-degenerative occlusion (IoU $\leq 0.67$), our method has consistent improvement over \INTER. 
In the high IoU regime ($>0.67$), the improvement levels off, which is expected since the second hand is almost entirely invisible and the problem is no longer caused by ambiguities. Here, it would be extremely challenging to reliably estimate the correct pose from a single image.

\begin{figure*}
\centering
  \includegraphics[width=0.8\linewidth]{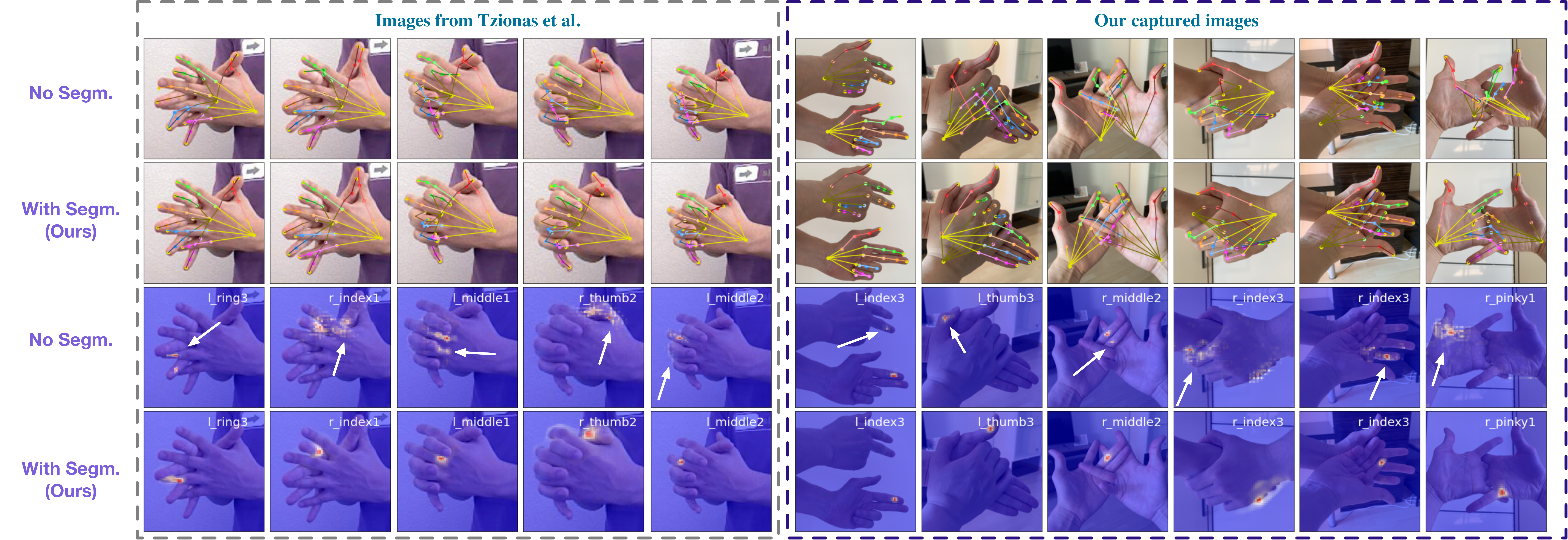}
  \caption{\textbf{Qualitative comparison of baseline and ours on images from\ccite{tzionas2016capturing} and newly captured in-the-wild images.} Brighter colors indicate left hand. The skeleton notation is provided in \suppl.
  Best viewed in color and zoomed in.}
  \label{fig: generalize}
  \vspace{-4mm}
\end{figure*}

\myparagraph{Generalization} \noindent
Our method addresses the ambiguities from the self-similarities between strongly interacting hands. To show generalization, we evaluate the baseline and ours on the interacting hand subset of\ccite{tzionas2016capturing}. The results show consistent improvements in MPJPE (ours/baseline: 15.68mm/17.37mm). More details are in the \suppl. Figure \ref{fig: generalize} shows that the ambiguity issue is present outside of \INTER and that our method is visibly more robust. Note that we use\ccite{tzionas2016capturing} for \textit{evaluation only} and do not train on it. Since all models are trained on\ccite{interhand}, and its images have a characteristic background, the trained models expect the same background in a more general setting.
To this end, we use an online service\ccite{rmbg} for background removal on the images of\ccite{tzionas2016capturing} and superimpose the images onto a background of\ccite{interhand}. Similarly, to show how the models perform under different lighting conditions, we also capture in-the-wild images and the results are in \reffig{fig: generalize}.

\myparagraph{Full probabilities vs part-labels}
Existing body-pose methods\ccite{omran2018neural, zanfir2020weakly} decouple the image-to-pose problem into image-to-segm.~and segm.-to-pose.
They use class-label segmentation maps, while our method leverages the full segmentation probability distribution.
To compare the impact of the two, we trained two separate networks with the two different approaches using the architecture in \reffig{fig: baseline_sl}c. The MPJPE values in the validation sets are 20.01mm and 25.49mm for the probabilistic map and the class-label map respectively, which shows that the probabilistic map preserves more information for downstream pose estimation.

\myparagraph{End-to-end training} \noindent
To show the effect of end-to-end training compared to training the image-to-segm.~and segm.-to-pose models separately\ccite{omran2018neural}, we compare two networks with the \textit{same} architecture (\reffig{fig: baseline_sl}c) but trained differently.
For the \emph{first} network, we trained its image-to-segm.~module for 25 epochs and then trained its segm.-to-pose module for 25 epochs while keeping the image-to-segm.~module fixed.
For the \emph{second} network, we trained the entire image-to-pose network for only 25 epochs to ensure the image-to-segm.~modules of both networks were trained with the same number of epochs. 
With/without end-to-end training yields an MPJPE of 20.01mm/23.52mm.

\subsection{Ablation study}
Here we provide insights on how, when and why our proposed method (\reffig{fig: pipeline}) improves over the baseline (\reffig{fig: baseline_sl}a).
Particularly, we examine the impact of the bone loss $\mathcal{L}_{b}$ (BL), the part segmentation loss $\mathcal{L}_{s}$ (SL), and the segmentation features $\M{S}$ (SF) for hand pose estimation. 

From \reftab{tab: contri_ih}, comparing the baseline with or without the bone loss, using the same architecture (\reffig{fig: baseline_sl}a), we see a 1.17mm improvement of 3D pose estimation for interacting hands on the test set.
Inspired by multi-task learning\ccite{lu202012}, we also study whether predicting part segmentation regularizes hand pose estimation. Thus, we train a pose estimator with an additional head to predict the part segmentation (see \reffig{fig: baseline_sl}b) but do not use segmentation for pose estimation.
The result shows that the segmentation loss improves pose estimation over the baseline with bone loss by 0.49mm on the test set.
For the relative root position, the segmentation loss dramatically improves MRRPE by 5.97mm on the test set over the baseline with bone loss. The reason is that, to satisfy $\mathcal{L}_s$, the backbone has to distinguish the roots of the two hands and the distinction reduces appearance ambiguity, resulting in better root localization.

Finally, in addition to the bone loss and the segmentation loss, our model (see \reffig{fig: pipeline}) makes use of the segmentation and visual features for pose estimation. 
Compared to the baseline with bone loss, our model reduces MPJPE by 1.38mm/0.99mm on the val/test sets. Further, we improve MRRPE by 5.97 mm on the test set.
We also trained a network with the same architecture but not supervised by the segmentation loss (Baseline + BL + SF\textsuperscript{*}). Since the performance of Baseline + BL is similar to Baseline + BL + SF\textsuperscript{*}, our improvement is not from the additional parameters.

\begin{table}[]
\centering
\resizebox{\columnwidth}{!}{%
\begin{tabular}{lcccc}
\toprule
Ablation Study     & MPJPE Val & MRRPE Val & MPJPE Test & MRRPE Test \\\hline
Baseline         & 21.91  & 36.73 & 18.71  & 34.05 \\
Baseline + BL      & 20.43  & 38.84 & 17.54  & 37.36 \\
Baseline + BL + SL  & 19.72 & 34.51 & 17.05  & 32.10 \\
Baseline + BL + SF\textsuperscript{*}   & 20.31   & 40.09 & 17.60 & 38.20 \\
Baseline + BL + SF (ours)   & \textbf{19.05}   & \textbf{34.14} & \textbf{16.55}  & \textbf{31.39} \\
\bottomrule
\end{tabular}
}
\caption{Effects of bone loss (BL), segm. loss (SL), and segm. features (SF). The symbol * denotes not using segm. supervision.}
\vspace{-4mm}
\label{tab: contri_ih}
\end{table}

\section{Discussion}
Our insight is that modelling part segmentation helps address ambiguities arising from self-similarity between hands. While our approach resembles 3D body-pose methods \ccite{omran2018neural, zanfir2020weakly} in leveraging part segmentation, the key difference is that we incorporate the segmentation task in an end-to-end fashion, leading to more informative features and an improvement in the main task. 
Particularly, we pass the unnormalized segmentation probability (\ie, logits) to the pose estimator, preserving the uncertainty for the downstream task. In contrast, \ccite{omran2018neural, zanfir2020weakly} take quantized features (\ie, class labels).
Our method also enables fully-differentiable end-to-end learning of hand pose estimation in conjunction with segmentation. We show that our end-to-end multi-task setup performs better compared to separate training of tasks and class-label inputs as in \ccite{omran2018neural, zanfir2020weakly}.

\section{Conclusion}
We introduce a method for interacting 3D hand pose estimation that explicitly addresses self-similarity between joints. Our method consists of two interwoven branches that process an input image into a per-pixel part segmentation mask and a visual feature volume. The part segmentation mask provides semantic features for visually similar hand regions while the visual feature volume provides rich visual cues for accurate pose estimation.
Our proposed method achieves SOTA performance on \INTERHAND. Our ablation studies show the efficacy of our method and provide insights into how the modeling of pixel ownership addresses self-ambiguity in interacting hand pose estimation.

\myparagraph{Acknowledgement}
We thank Korrawe Karunratanakul, Emre Aksan, and Dimitrios Tzionas for their feedback.

\myparagraph{Disclosure}
MJB has received research gift funds from Adobe, Intel, Nvidia, Facebook, and Amazon. While MJB is a part-time employee of Amazon, his research was performed solely at, and funded solely by, Max Planck. MJB has financial interests in Amazon, Datagen Technologies, and Meshcapade GmbH.

{\small
\bibliographystyle{ieee_fullname}
\bibliography{egbib}
}

\title{\TITLE\\ **Appendix**}
\author{Zicong Fan$^{1,2}$ \quad Adrian Spurr$^{1}$ \quad Muhammed Kocabas$^{1,2}$ \quad Siyu Tang$^{1}$ \\Michael J. Black$^{2}$ \quad Otmar Hilliges$^{1}$\vspace{0.1cm} \\
 $^1$ETH Z{\"u}rich, Switzerland \quad
 $^2$Max Planck Institute for Intelligent Systems, T{\"u}bingen
}

\twocolumn[{%
\renewcommand\twocolumn[1][]{#1}%
\maketitle
\thispagestyle{empty}

\begin{center}
  \newcommand{\teaserwidth}{\textwidth}
  \vspace{-0.7cm}
  \centerline{\includegraphics[width=1.0\linewidth]{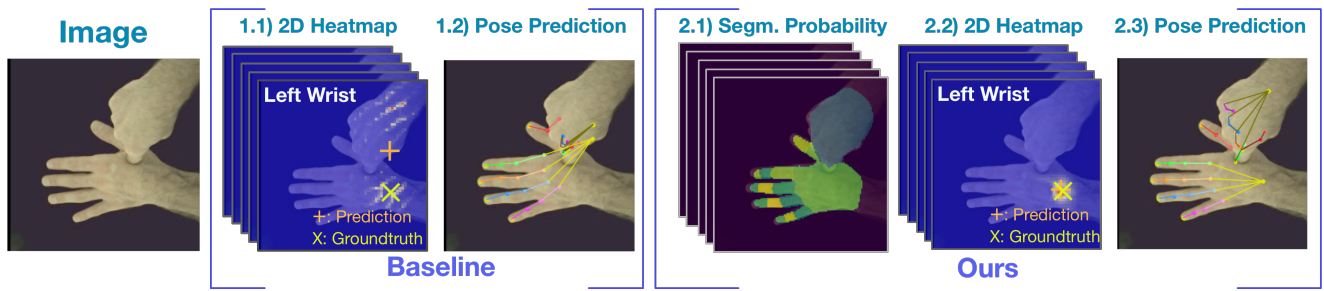}}
    \captionof{figure}{
    When estimating the 3D pose of interacting hands, state-of-the-art methods struggle to disambiguate the appearance of the two hands and their parts. In this example, significant uncertainty between the left and right wrist arises (1.1), resulting in erroneous pose estimation (1.2). Our model, DIGIT, reduces the ambiguity by predicting and leveraging a probabilistic part segmentation volume (2.1) to produce reliable pose estimates even when the two hands are in direct contact and under significant occlusion (2.2, 2.3).}
    \label{matfig: teaser}
\end{center}%
}]

\begin{figure*}[t]
\centering
  \includegraphics[width=0.8\linewidth]{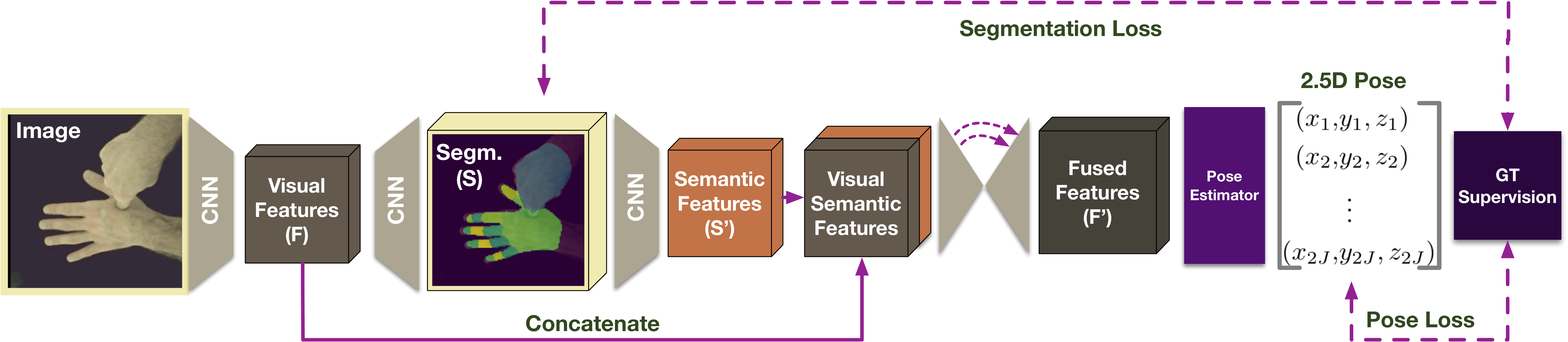}
  \caption{\textbf{An illustration of our hand pose estimation model.} Given an image, our method extracts visual features ($\M{F}$) and predict part segmentation probability volume ($\M{S}$). The segmentation volume is projected into latent semantic features ($\M{S'}$). The visual features ($\M{F}$) and the semantic features ($\M{S'}$) are fused across multiple scales and are used for interacting hand pose estimation (illustrated in \reffig{matfig: pose_est}).}
  \label{matfig: pipeline}
\end{figure*}

\begin{figure}[]
\centering
  \includegraphics[width=1.0\linewidth]{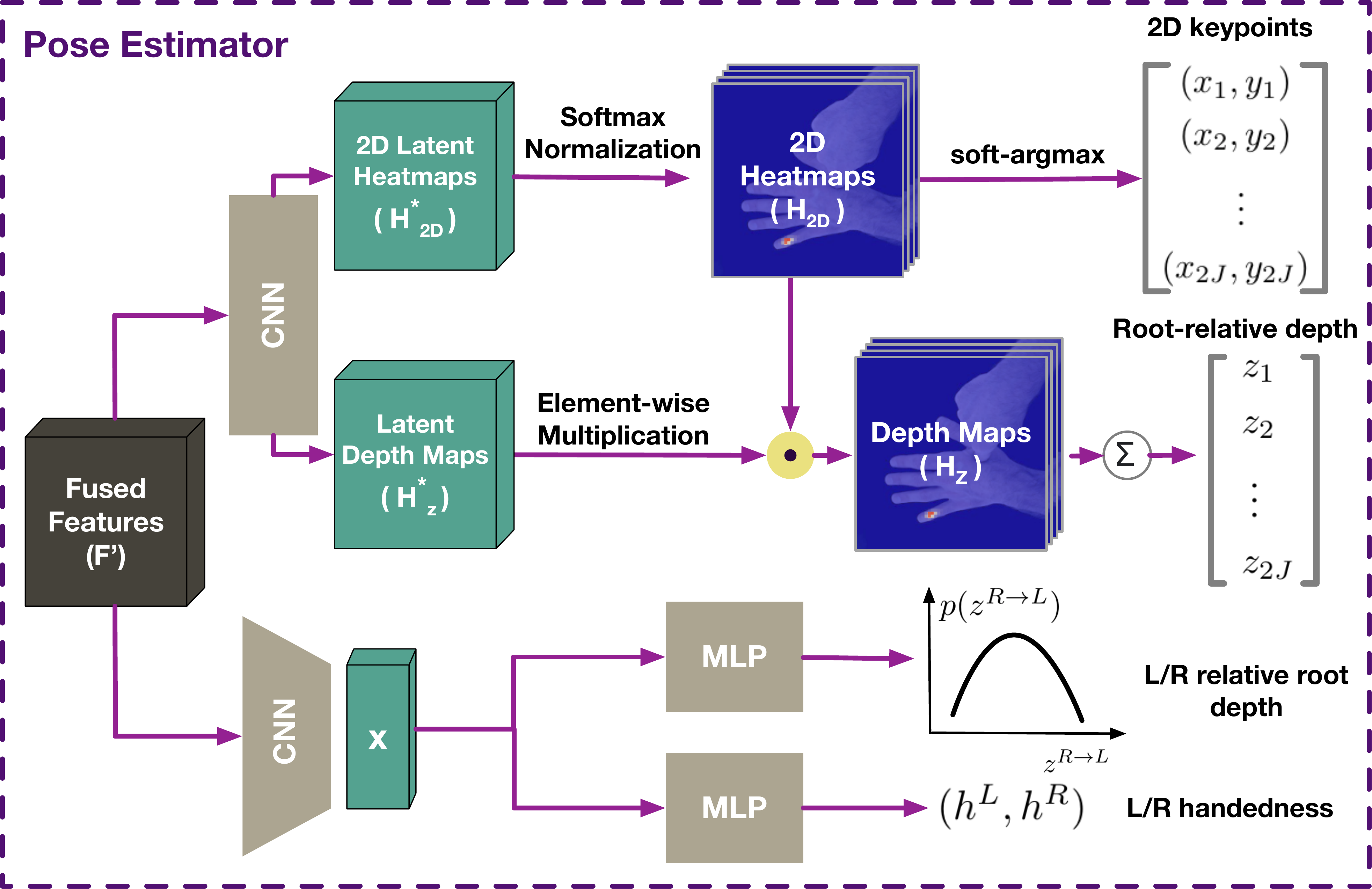}
  \caption{\textbf{Our interacting hand pose estimator}}
  \label{matfig: pose_est}
\end{figure}
\begin{figure}[]
\centering
  \includegraphics[width=1.0\linewidth]{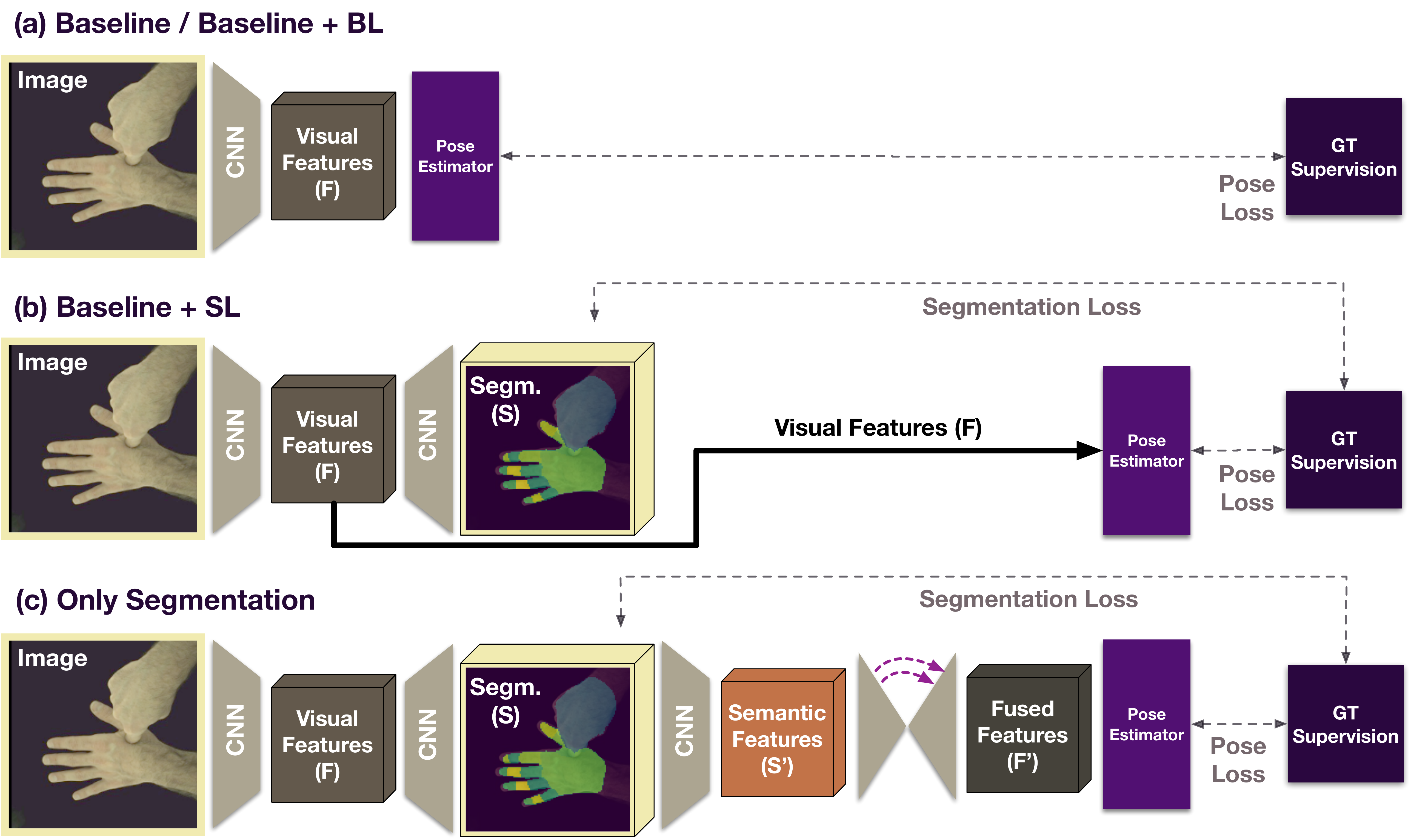}
  \caption{\textbf{Pose estimators for the ablation studies:} (a) without segmentation, (b) with segmentation but only for applying a loss, (c)  segmentation for pose estimation (no visual features).}
  \label{matfig: baseline_sl}
\end{figure}

\textit{The \suppl.  includes this document and \textbf{a short video}.} \\

In natural interaction, our hands often occlude or are in contact with one and another. However, due to the homogeneous appearances of hands, estimating 3D hand poses from a single RGB image is difficult. This self-similarity in appearances is problematic during hand-to-hand interaction, which often includes complex occlusion patterns. 
Our experiments in the main paper show that the state-of-the-art interacting hand pose estimators do not have a coping mechanism for such ambiguity. To go beyond the limits of existing methods, we propose a simple yet effective network to estimate interacting 3D hand poses by modeling per-pixel ownership via probabilistic part segmentation.

To complement the experiments in our main paper, we provide additional details.
We first provide more qualitative examples showcasing the appearance ambiguity problem, the limits of the \INTERHAND model, the benefit of modeling part segmentation to address the ambiguity, and the 3D predictions from our model.  
In the next section, we analyze of our method, illustrating the relationship between segmentation prediction and pose estimation quality, the impact of interaction and occlusion on segmentation prediction, and the generalization of our method.
Lastly, we present the implementation details in the final section.

\section{Qualitative analysis}\label{sec: more_examples}

To demonstrate the appearance ambiguity problem and its impact on the \INTERHAND model, we evaluate the official release of a pre-trained model\ccite{moongithub}. Since the model uses a volumetric heatmap representation to encode the 2.5D location of individual joints for an input image, we sum across the volumetric heatmap for each joint along the depth dimension to obtain a 2D heatmap, representing the location of 2D keypoints in the image space. \reffig{matfig: suppl_bimodal_ih} shows examples of such 2D heatmaps. 
Note that the lighting of the input images is adjusted for visualization purposes but the original image lighting is used for the model input.

\begin{figure}
\centering
  \includegraphics[width=0.6\linewidth]{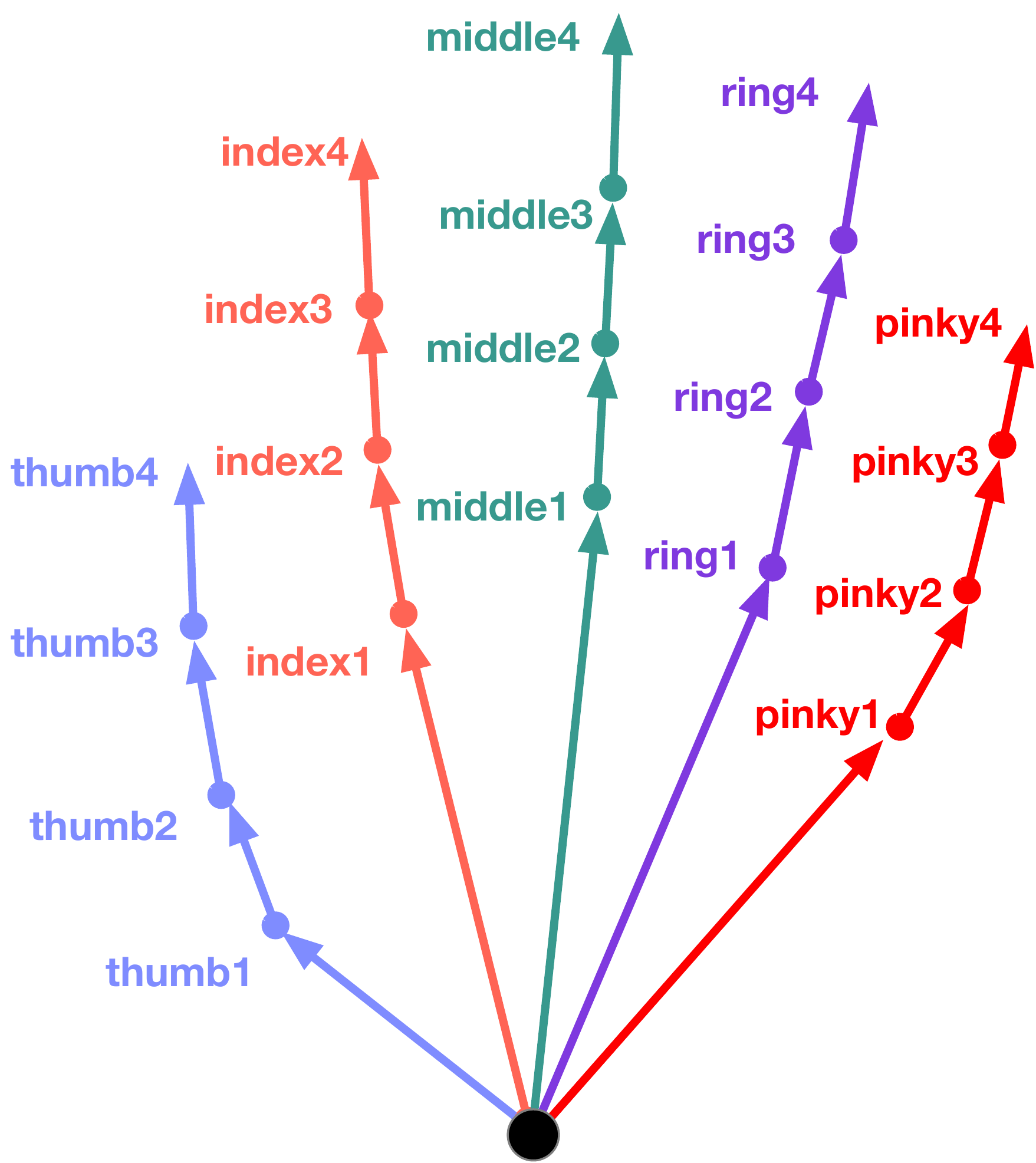}
  \caption{\textbf{Skeleton notation for each joint.} }
  \label{matfig: suppl_notation}
\end{figure}

We show the 2D heatmaps of individual joints as opposed to all 42 joints to avoid a cluttered visualization. As an example, the top-left corner shows an image of a single hand. The 2D heatmap of the ``ring3" joint (see \reffig{matfig: suppl_notation}) is shown below the input image.
We observe multiple modes in the predicted 2D heatmap in different parts of the hand. This indicates that the model has difficulties in resolving ambiguities. 
Note that since the \INTERHAND model uses an argmax operation over the 2D heatmap to determine the 2D keypoint location, the prediction is at the mode with the highest response, shown as a plus sign in the figure. We use the cross sign to indicate the groundtruth location.

To understand how the modeling of part segmentation helps to address appearance ambiguity, we provide qualitative examples of models with or without using segmentation in \reffig{matfig: suppl_qualitative_segm}. 
The model with segmentation is illustrated in \reffig{matfig: pipeline} while the model without segmentation is in \reffig{matfig: baseline_sl}a.
We show the 2D heatmaps of individual joints to avoid clutter in the visualization.
For instance, the middle example at the bottom row in \reffig{matfig: suppl_qualitative_segm} shows that, without part segmentation, the network struggles to distinguish the left and the right wrist, indicated by the two modes on the hands. In contrast, by predicting segmentation, which distinguishes left and right (different colors on the predicted segmentation map), our 2D heatmap has a single mode, concentrated at the wrist area of the left hand. 
Similarly, since we predict part segmentation, the ambiguity between parts can be addressed in the same way.
A concentrated heatmap is often observed for visible joints in our model. However, when a joint is occluded (the fourth example at the bottom of \reffig{matfig: suppl_qualitative_segm}), our model leverages context from other visible joints, shown as the spread of weights on the heatmap.
Since we use soft-argmax\ccite{luvizon2019human} to convert 2D heatmaps to 2D keypoints for full differentiability, the predicted 2D keypoint of each joint is at the weighted-average location according to the 2D heatmap (see \refeq{mateq: softargmax2d}).
We use the plus sign and the cross sign to indicate the predicted 2D location and the groundtruth.  The name of each joint is shown at the top of each example (see \reffig{matfig: suppl_notation} for skeleton notation).

Finally, \reffig{matfig: suppl_qualitative} provides 3D predictions from our model, including the input image, the 2D keypoints, the part segmentation prediction, and the 3D predictions in two different views. The groundtruth for each type of prediction is shown for a side-by-side comparison. To visualize left and right, we use darker bone colors for right-hand bones and brighter for the left hand. Similarly, darker segmentation masks are for the right hand and brighter masks are for the left hand. The figure shows that despite the very similar appearance between hands and their parts, 
our segmentation network produces reasonable predictions in distinguishing different parts, shown by having the same color in the predicted and the groundtruth part segmentation.

\section{Analysis of our method}\label{sec: analysis}
Here, we provide further analysis of our method. For example, we discuss the relationship between 3D pose estimation performance and segmentation prediction quality. We also study the quality of segmentation prediction to the amount of interaction, and the generalization of our method. Finally, we show the computational footprint of the baseline network, \INTER, and our model.

\myparagraph{Pose estimation \vs segmentation quality}
In \reftab{mattab: miou_mpjpe}, we study the relationship between segmentation quality and pose estimation performance on the validation set.
In particular, we evaluate our model (see \reffig{matfig: pipeline}) and compute the mean Intersection over Union (mIoU) between part segmentation prediction and the groundtruth for each image, which measures how similar the predicted segmentation is to the groundtruth. A higher mIoU value indicates higher similarity. 
We partition the images according to different mIoU ranges (top row of the table) and compare our model with the baseline (see \reffig{matfig: baseline_sl}a). 
Both models use the same training procedure for a fair comparison. 
We observe that MPJPE (mm) and mIoU are negatively correlated and the standard deviation of MPJPE decreases for images with higher mIoU. 
Compared to the baseline, overall our model has a lower mean and standard deviation in the MPJPE metric.
When mIoU is in $[0.0, 0.2)$, both models have high MPJPE values.
This mIoU range includes images with degenerate occlusion, resulting in both low-quality segmentation and noisy 3D pose prediction ($\pm 36\sim 37$mm).

\begin{table}[b]
\centering
\resizebox{\columnwidth}{!}{%
\begin{tabular}{lccccc}
\toprule
mIoU range & {[}0.0, 0.2) & {[}0.2, 0.4) & {[}0.4, 0.6) & {[}0.6, 0.8) & {[}0.8, 1.0{]} \\\hline
W/O segm.  & \textbf{42.0} $\pm$ 37.0 & 22.1 $\pm$ 9.5  & 18.7 $\pm$ 7.1  & 15.8 $\pm$ 5.3  & 12.6 $\pm$ 3.8    \\
W/ segm. & 43.3 $\pm$ \textbf{36.1} & \textbf{20.9}$\pm$ \textbf{8.2}  & \textbf{17.1} $\pm$ \textbf{6.1}  & \textbf{14.6} $\pm$ \textbf{4.7}  & \textbf{11.8} $\pm$ \textbf{3.6}    \\\hline
Difference & -1.3 $\pm$ 0.9  & 1.2 $\pm$ 1.3   & 1.6 $\pm$ 1.0   & 1.2 $\pm$ 0.6   & 0.8 $\pm$ 0.2 \\
\bottomrule
\end{tabular}}
\caption{\textbf{Comparing models with or without segmentation.} Each entry denotes the MPJPE (mm) and its standard deviation. To measure the segmentation quality of our model, we compute the mIoU between the part segmentation prediction and the groundtruth. The mIoU is then used to partition the images and compare the MPJPE of the two models in each range.
}
\label{mattab: miou_mpjpe}
\end{table}

\begin{figure}[h!]
\centering
  \includegraphics[width=1.0\linewidth]{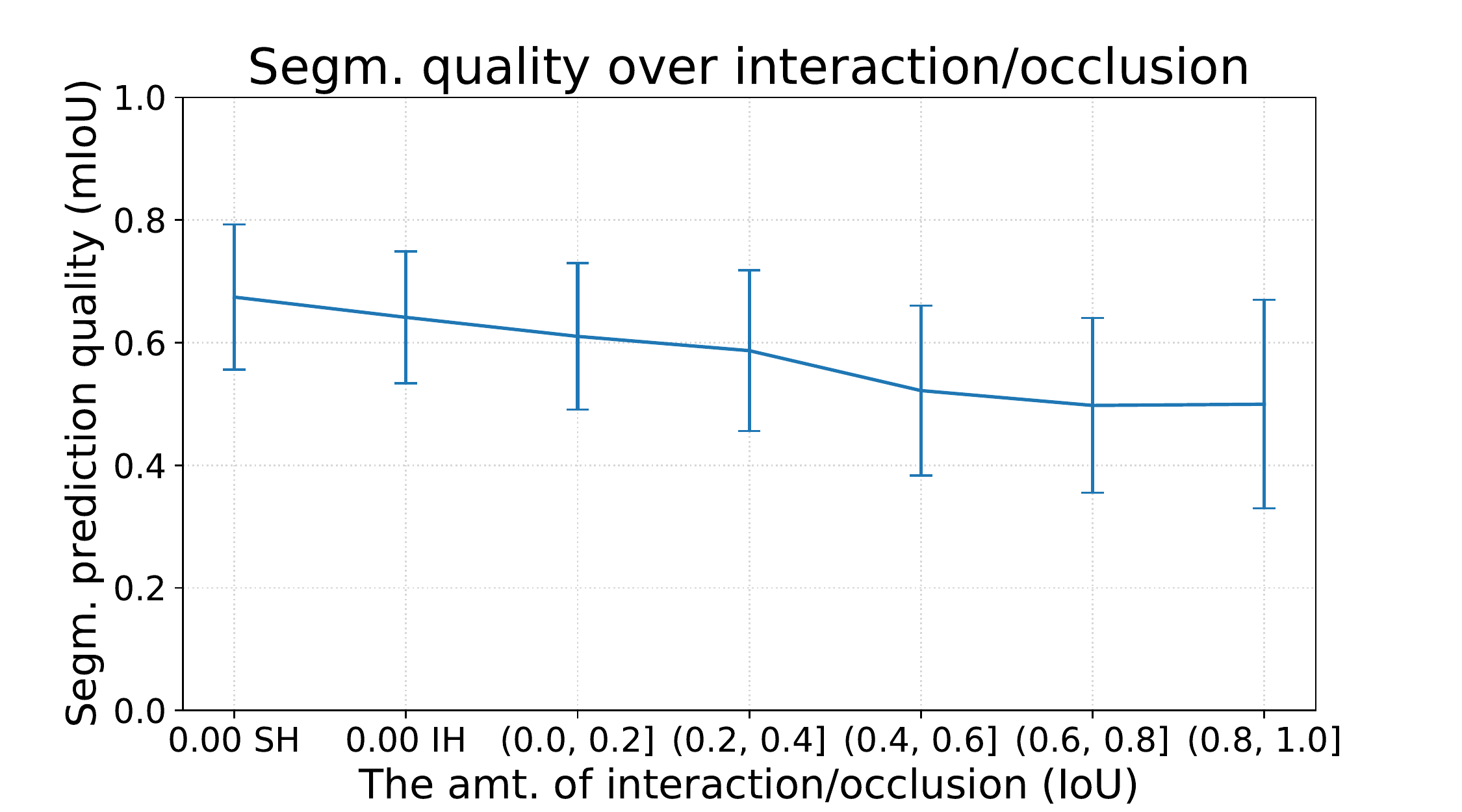}
  \caption{\textbf{Segmentation prediction quality for different amounts of interaction and occlusion.} We use Intersection over Union (IoU) between the groundtruth left and right full hand masks to measure the amount of interaction and occlusion. The segmentation quality is measured by the mean IoU (mIoU) between the predicted segmentation and the groundtruth.}
  \label{matfig: miou_iou}
\end{figure}

\myparagraph{Segmentation quality \vs  interaction}
\reffig{matfig: miou_iou} shows the segmentation prediction quality of our model for images containing different amounts of interaction.
We use the Intersection over Union (IoU) between the groundtruth left and right full hand masks (not part segmentation) to measure the degree of interaction and occlusion in each image, shown on the horizontal axis. On the vertical axis, we measure the segmentation prediction quality by computing the mean IoU (mIoU) between the predicted part segmentation and the groundtruth. The higher mIoU indicates better segmentation prediction while the higher IoU indicates more interaction and occlusion in the image.
The figure shows that the segmentation quality decreases as interaction increases but plateaus at around 0.5. This indicates that part segmentation is still feasible during strong hand interaction.

\begin{table}[h]
\vspace{-1mm}
\centering
\resizebox{0.8\columnwidth}{!}{%
\begin{tabular}{cccc}
\toprule
     & \INTERHAND & Baseline    & Ours \\\hline

Val  & 66.80/91.95             & 66.32/92.71 & \textbf{63.12}/\textbf{90.55} \\
Test & 70.66/83.81             & 69.58/78.56 & \textbf{68.00}/\textbf{77.28}\\
\bottomrule
\end{tabular}}
\caption{\textbf{Absolute 3D pose errors without root alignment (mm) for single/interacting hand sequences.}}
\label{mattab: abs_err}
\end{table}

\myparagraph{Absolute 3D error}
Following the evaluation protocol of\ccite{interhand}, we only include root-relative errors in the main paper. However, as a reference, we provide the 3D absolute pose error without root alignment in \reftab{mattab: abs_err}. For a fair comparison, all methods use the same absolute root depth values predicted using RootNet\ccite{moon2019camera}, which are provided in\ccite{interhand}. Note that to measure 3D hand pose estimation performance, root-relative 3D errors (MPJPE) is more suitable than the absolute error because it is not dominated by the errors of root depth estimation from RootNet\ccite{moon2019camera}.

\myparagraph{Backbones}
To show that our method is backbone-agnostic, we trained the baseline and our method with ResNet50\ccite{he2016deep}. Our method consistently outperforms the baseline by 1.40mm/1.73mm for single/interacting hands on the validation set. 
The two networks are trained for 50 epochs with an initial learning rate of $10^{-3}$ and decay by a factor of 10 at epoch 40.

\myparagraph{Generalization}
Since images of\ccite{interhand} have a characteristic background and contain similar lighting on hands, we show that the ambiguities between interacting hands still exist on images with more natural lighting, and that our method addresses such ambiguities. Thus, we use the dataset from Tzionas \etal \ccite{tzionas2016capturing} for testing purposes and we \textit{do not train} our models on\ccite{tzionas2016capturing}. To our knowledge, it is the only published RGB dataset containing accurate annotations for strongly interacting hands with natural lighting. 

Since\ccite{interhand} is manually annotated while the groundtruth skeleton of \ccite{tzionas2016capturing} is from MANO\ccite{mano} fitting, the skeleton topologies of the two datasets are inconsistent. This makes evaluating the models on\ccite{tzionas2016capturing} difficult. To this end, following SMPLify\ccite{bogo2016keep}, for each hand, we use a small MLP to map the skeleton of\ccite{interhand} to MANO skeleton to compare with the groundtruth from\ccite{tzionas2016capturing}. The MLP takes the manual annotated skeleton of the left/right hand as inputs and regresses to its corresponding MANO skeleton. The regressor contains a 512-dimensional hidden layer connected by ReLU\ccite{relu}, and is trained using the training set of\ccite{interhand} until convergence. Two regressors are trained and used respectively for the left and the right hands. The regressors are not finetuned on\ccite{tzionas2016capturing}. To convert 2.5D to 3D joints, we use the groundtruth root depth values from the MANO skeleton.

\myparagraph{Single-hand metrics}
Although our paper is about addressing the ambiguities from the self-similarities between joints during hand-to-hand interaction, we provide the single-hand root-relative mean pose per joint error (MPJPE) for references (see \reftab{mattab: sota_sh}). 
T-tests on the MPJPE values reveal that all differences between ours and the baseline are statistically significant (all $p<10^{-4}$) for both val/test sets. Our improvement over \INTERHAND and the baseline is smaller than that of the interacting hand sequences due to fewer joints to cause the ambiguities. 

\begin{table}[H]
\centering
\resizebox{\columnwidth}{!}{%
\begin{tabular}{lcc}
\toprule
Methods & MPJPE Val  & MPJPE Test \\\hline
\INTERHAND      & 14.82 & 12.63 \\
Baseline & 14.64 & 12.32 \\
Ours & \textbf{13.54} & \textbf{11.32}\\\hline
\% in improvement over\ccite{interhand} & 8.64 & 10.37\\
\bottomrule
\end{tabular}
}
\caption{\textbf{Comparison with SOTAs on single-hand sequences on the initial release of\ccite{interhand}} The MPJPE metric measures the accuracy of interacting 3D hand pose estimation. Trained and evaluated on the initial release of \INTERHAND.}
\label{mattab: sota_sh}
\end{table}

\begin{table}[h]
\centering
\resizebox{\columnwidth}{!}{%
\begin{tabular}{lcccc}
\toprule
Methods & MPJPE Val  & MRRPE Val  & MPJPE Test  & MRRPE Test \\\hline
\INTERHAND      & 20.59 & 35.99 & 17.36                           & 34.49 \\
Baseline & 20.24 & 35.05 & 17.23 & 32.70                           \\
Ours & \textbf{18.28} & \textbf{32.21} & \textbf{15.57} & \textbf{30.51}\\\hline
\% in improvement over\ccite{interhand} & 11.22 & 10.50 & 10.31 &11.54\\
\bottomrule
\end{tabular}
}
\caption{\textbf{Comparison with SOTAs on interacting-hand sequences on the initial release of\ccite{interhand}.} The MPJPE metric measures the accuracy of interacting 3D hand pose estimation while the MRRPE metric shows the performance of the relative root position between the two hands. Trained and evaluated on the initial release of \INTERHAND. }
\label{mattab: sota_ih}
\end{table}

\myparagraph{SOTA comparison on the initial release of\ccite{interhand}}
We show our results compared with the state-of-the-art methods trained and evaluated on the initial release of \INTERHAND in \reftab{mattab: sota_ih}.

\myparagraph{Impact of prob. formulation and segm. granularity} We include an experiment of a network (see Fig.~6c of the main paper) predicting masks for the left/right hands instead of probabilistic part segmentation (see Table \ref{mattab: lr_masks} below). Our probabilistic part segmentation is more accurate, indicated by the lower MPJPE error compared to only left/right-hand segmentation (see ``LR masks"). We also show how the segmentation granularity affects pose estimation. The row ``Prob.~LR segm." is from a model that predicts left/right segmentation using our probabilistic formulation.

\begin{table}[h]
\centering
\resizebox{0.8\columnwidth}{!}{%
\begin{tabular}{lcccc}
\toprule
Ablation Study     & MPJPE Val  & MPJPE Test \\\hline
Prob.~part segm. (ours)      & \textbf{20.01}  & \textbf{17.23} \\
Prob.~LR segm.  & 21.19 & 18.40 \\
LR masks  & 36.05 & 31.46 \\
\bottomrule
\end{tabular}
}
\caption{Impacts of prob. formulation and segm. granularity}
\label{mattab: lr_masks}
\end{table}

\myparagraph{Segmentation resolution} 
Our choice of segmentation resolution trades-off accuracy and efficiency. In particular, going from 128 to 256 dimensions does not differ significantly in performance ($18.51$mm vs. $18.43$mm in the validation set) but increases the training time by 60\%.

\myparagraph{Computation requirement}
Since the \INTERHAND model has a high memory requirement, we extend the pose estimator by Iqbal \etal\ccite{iqbal2018hand} for modelling interacting hand pose. \reftab{mattab: compute} shows the number of parameters of the \INTERHAND model, our baseline extended from\ccite{iqbal2018hand} (shown in \reffig{matfig: baseline_sl}a), and our final model (\reffig{matfig: pipeline}).
The baseline has significantly fewer parameters and lower memory requirement than\ccite{interhand}. This justifies our need to use a custom pose estimator. Our final model has fewer parameters than\ccite{interhand} while requiring $1.2$ Gb more memory due to modeling part segmentation. The baseline, our method, and\ccite{interhand} run at 26, 24, and 96 FPS on a Titan RTX. Training our method for 30 epochs takes 5 to 7 days due to the dataset size on a single Quadro RTX 6000 GPU.

\begin{table}
\centering
\begin{tabular}{lccc}
\toprule
         & \# Parameters & Memory \\\hline
\INTERHAND & 47.3 M & 13.8 Gb\\
Baseline &  35.0 M & 8.5 Gb\\
Ours     &  40.0 M & 15.0 Gb\\
\bottomrule
\end{tabular}
\caption{\textbf{Number of parameters and memory requirement for training the two baseline models and our final model.} Our baseline is shown in \reffig{matfig: baseline_sl}a. The memory requirement is measured on a Quadro RTX 6000 GPU with a batch size of 32.}
\label{mattab: compute}
\end{table}

\section{Implementation details}\label{sec: details}
In this section, we provide implementation details of our pipeline, our custom UNet (see \reffig{matfig: pipeline}), and our pose estimator (see \reffig{matfig: pose_est}).
We use $W\times H\times D$ to denote the width, height, and channel dimension for all feature maps.

\myparagraph{Our pipeline} 
\reffig{matfig: pipeline} shows the architecture of our proposed model. Given an image, we use a HRNet-W32\ccite{sun2019deep} CNN backbone to obtain visual features $\M{F \in \R^{W_F\times H_F\times D_F}}$ where $W_F$, $H_F$ and $D_F$ are 64, 64, 32 respectively. The visual features are used to estimate the part segmentation in the form of probabilistic segmentation logits $\M{S}\in \R^{W_S\times H_S\times C}$ where $W_S=H_S=128$ and $C=33$. The quantity $C$ is the number of classes including the background. 
\reftab{mattab: supplsegm} illustrates the sequence of operations applied on the visual features $\M{F}$ to obtain the segmentation logits $\M{S}$. The network first upsamples $\M{F}$ by a factor of 2 using bilinear interpolation. It is then processed by a series of 2D convolution (with a kernel width of 3 and zero paddings of 1), 2D batch normalization, and rectified linear unit\ccite{relu} layers. The input and output channel dimensions of each layer are shown in the brackets.

\begin{table}[]
\begin{tabular}{cl}
\toprule
Nr. & \multicolumn{1}{c}{Layers/Blocks} \\\hline
0   & Upsample(scale=2, mode=bilinear) \\
1   &Conv2d(32, 16, kw=3, pad=1), BN2d(16), ReLU()\\
2   &Conv2d(16, 64, kw=3, pad=1), BN2d(64), ReLU()\\
3   &Conv2d(64, 33, kw=3, pad=1), BN2d(33), ReLU()\\
4   &Conv2d(33, 33, kw=3, pad=1)\\
\bottomrule
\end{tabular}
\caption{Details of the part segmentation network.}
\label{mattab: supplsegm}
\end{table}

Since the segmentation $\M{S}$ has a higher spatial resolution than the visual features $\M{F}$, to match the dimension of $\M{S}$ to $\M{F}$ while preserving the higher resolution semantic information, we convolve  $\M{S}$ with two 2D convolution layers whose output channel dimensions are 64 and 512. The convolved feature map is then downsampled spatially by a factor of 2 using max pooling, resulting in semantic features $\M{S'}\in \R^{W_F\times H_F\times D_S}$ where $D_S=512$. To introduce non-linearity, we use the ReLU\ccite{relu} activation function between the two 2D convolution layers and the max-pooling layer.

To leverage both visual features $\M{F}$, containing rich visual cues for accurate pose estimation, and semantic features $\M{S'}$, containing distinctive semantic information for addressing the appearance ambiguity problem, we concatenate $\M{F}$ and $\M{S'}$ along the channel dimension to a feature map with the dimensions $W_F\times H_F\times (D_F+D_S)$. We then use a UNet-like architecture to fuse the two feature maps across multiple scales in order to leverage global context. 
Finally, the fused features $\M{F'}$ are used to estimate the hand pose.

\begin{figure}
\centering
  \includegraphics[width=0.8\linewidth]{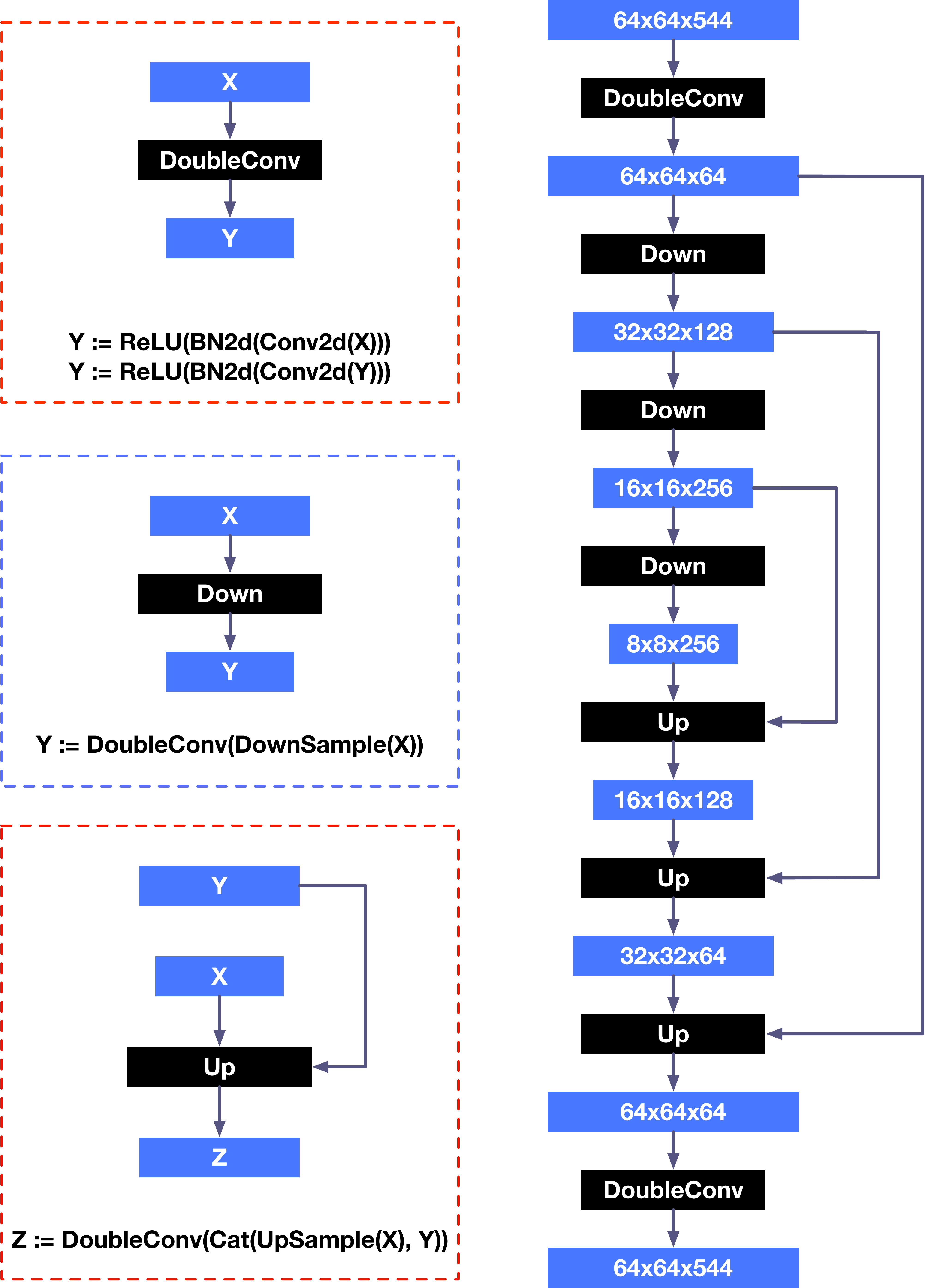}
  \caption{\textbf{Our UNet-like fusion network.} On the left, we define the notation for three major operations. On the right, the arrows define the input and output relationship between the feature maps and the operations.}
  \label{matfig: suppl_unet}
\end{figure}

\begin{figure*}
\centering
  \includegraphics[width=1.0\linewidth]{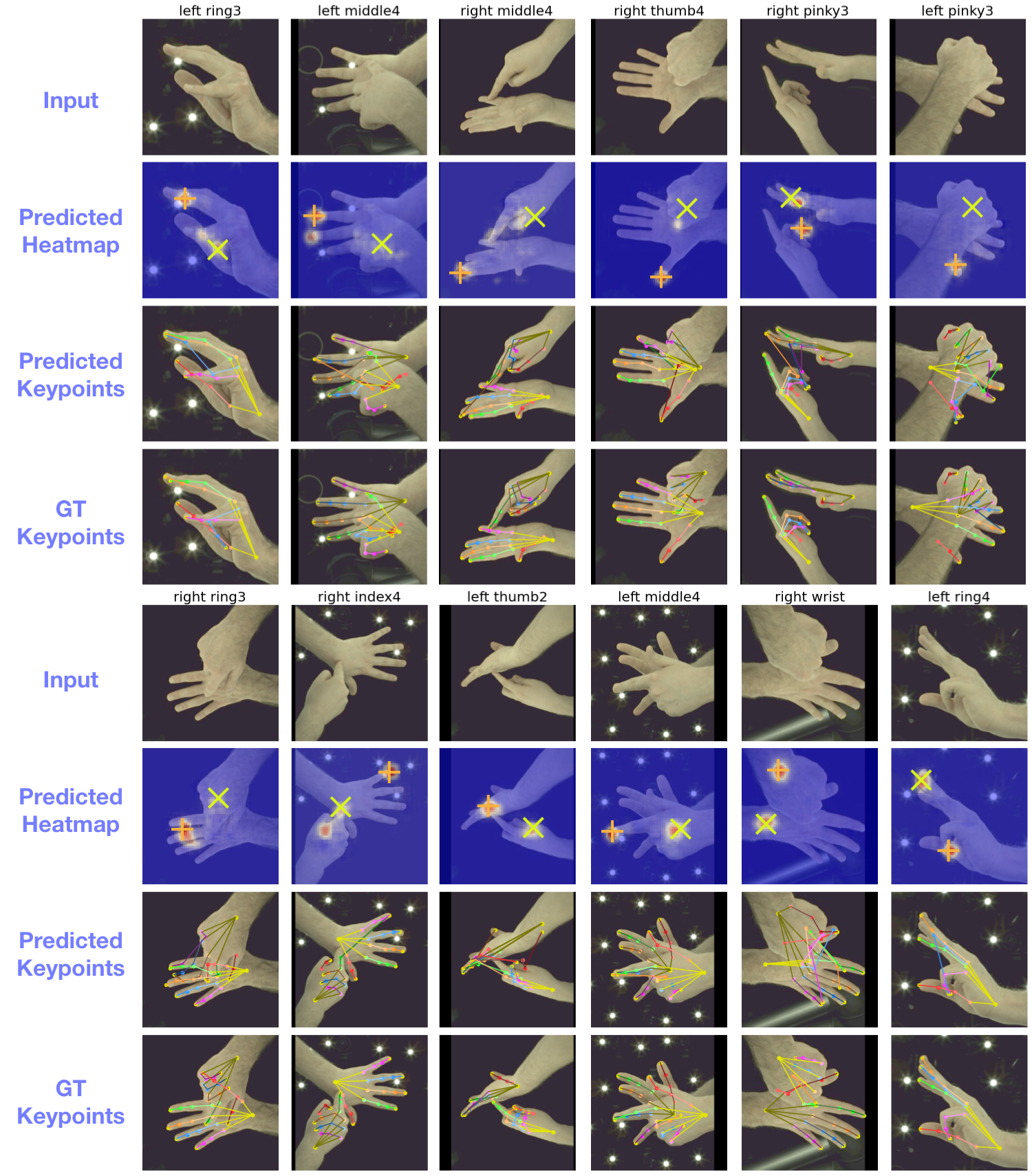}
  \caption{\textbf{Bi-modal and dispersed heatmap behavior in \INTER.} We show the input image, the 2D heatmap of the joints of interest (the joint name at the top), the groundtruth keypoints, and the predicted keypoints. In the 2D heatmap, the plus sign and the cross sign denote the predicted and groundtruth position for a particular joint, respectively. The naming convention of joints can be found in \reffig{matfig: suppl_notation}. Best viewed in color and zoomed in. }
  \label{matfig: suppl_bimodal_ih}
\end{figure*}

\begin{figure*}
\centering
  \includegraphics[width=1.0\linewidth]{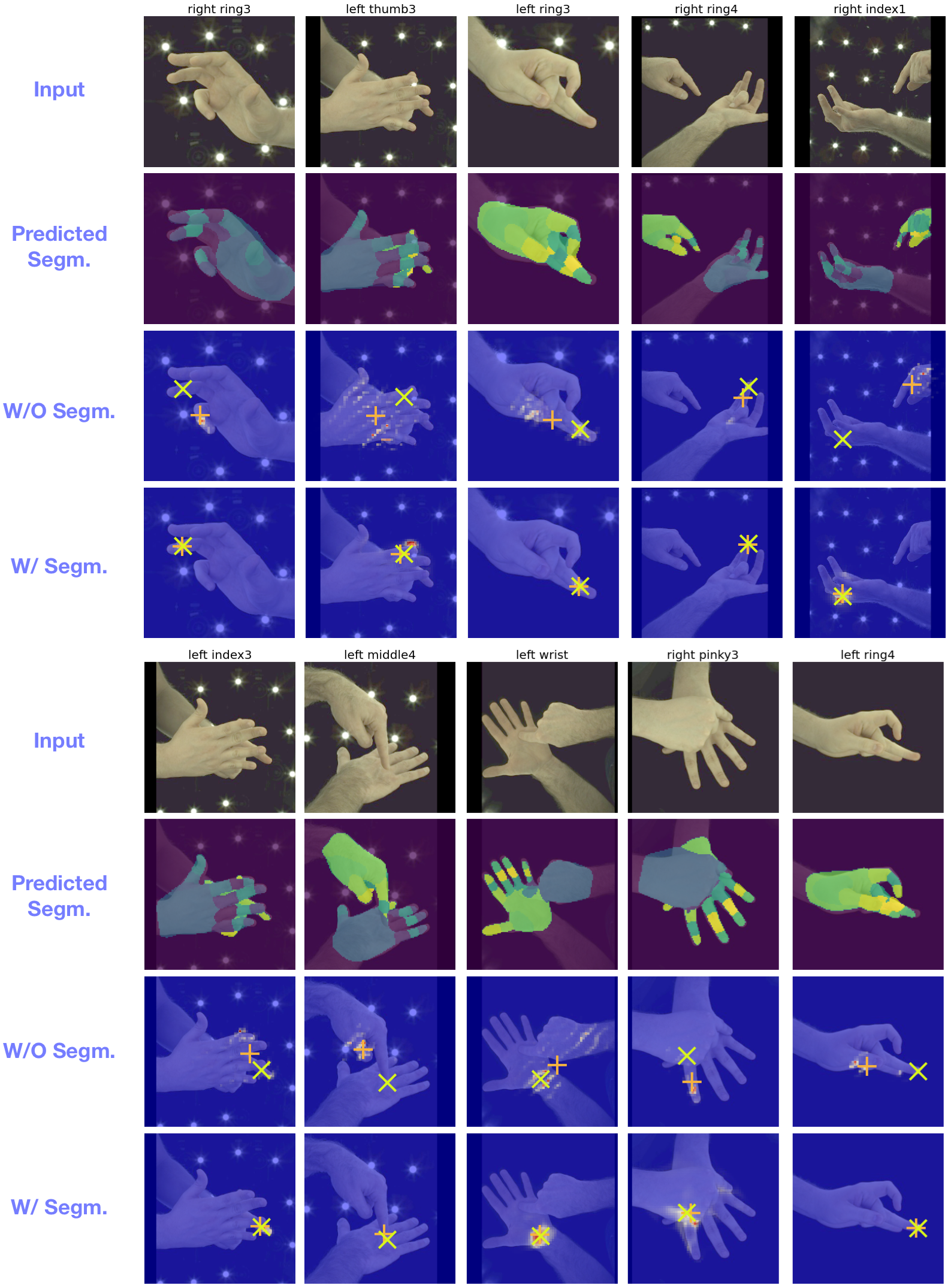}
  \caption{\textbf{Comparing heatmaps with or without using segmentation.} The darker
  segmentation masks denote right-hand predictions. The plus sign and the cross sign show the predicted and groundtruth position for a particular joint. Best viewed in color and zoomed in. }
  \label{matfig: suppl_qualitative_segm}
\end{figure*}

\begin{figure*}
\centering
  \includegraphics[width=1.0\linewidth]{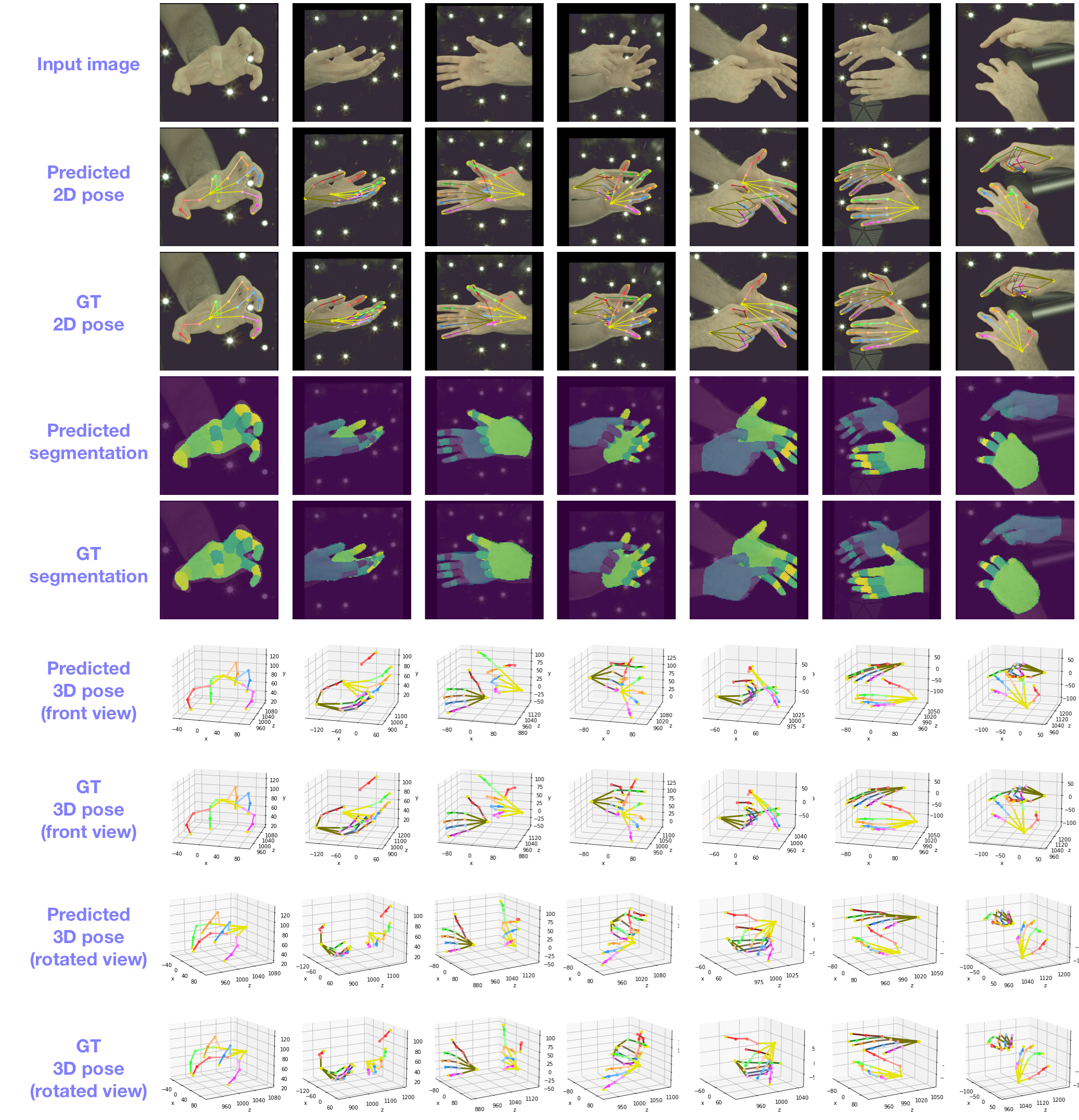}
  \caption{\textbf{Qualitative results of our model.} We use darker bones and segmentation colors to denote predictions for the right hands. 
  Best viewed in color and zoomed in. }
  \label{matfig: suppl_qualitative}
\end{figure*}

\myparagraph{The fusion network}
Given the feature map with the dimensions $W_F\times H_F\times (D_F+D_S)$, obtained by concatenating the visual features $\M{F}$ and the semantic features $\M{S'}$, we use a UNet-like architecture to fuse the two feature maps, which results in a fused feature map $\M{F'}$. The right hand side of \reffig{matfig: suppl_unet} illustrates the fusion network (4.6M parameters). The network consists of three types of operations: \texttt{DoubleConv}, \texttt{Down}, and \texttt{Up}, defined on the left side of the figure.
In particular, we use \texttt{DoubleConv} in the figure to denote applying the sequence of operations twice: a 2D convolution, a 2D batch normalization\ccite{batchnorm}, and a ReLU\ccite{relu} operation. 
The 2D convolution layers use $3\times 3$ kernel with a zero padding of 1 pixel to ensure the same input and output spatial dimensions. 
We use \texttt{Down} to denote a bilinear downsampling procedure followed by a \texttt{DoubleConv}. The downsampling procedure reduces the spatial dimensions of the input by a factor of 2. Finally, we use \texttt{Up} to denote an upsampling operation of a feature map with lower spatial dimensions (shown as $X$ in the smaller figure at the bottom-left) and a concatenation of the upsampled $X$ with the feature map $Y$ that matches the same dimension. 
After the concatenation, a \texttt{DoubleConv} is used to produce the feature map $Z$.
We also provide the pseudocode for the operations below each diagram for clarity.
The input and output feature maps of the three major operations are indicated by the arrow directions in the network.

\begin{table}[]
\begin{tabular}{cl}
\toprule
Nr. & \multicolumn{1}{c}{Layers/Blocks} \\\hline
0   &Conv2d(544, 64, kw=3, pad=1), ReLU()\\
1   &Conv2d(64, 128, kw=3, pad=1), ReLU()\\
2   &MaxPool2d(kw=2, stride=2)\\
3   &Conv2d(128, 256, kw=3, pad=1), ReLU()\\
4   &Conv2d(256, 512, kw=3, pad=1), ReLU()\\
5   &MaxPool2d(kw=2, stride=2)\\
6   &Conv2d(512, 512, kw=3, pad=1), ReLU()\\
7   &Conv2d(512, 512, kw=3, pad=1), ReLU()\\
8   &MaxPool2d(kw=2, stride=2), Mean2d()\\
\bottomrule
\end{tabular}
\caption{Converting fused features $\M{F'}$ to vector $\M{x}$ for handedness and relative root depth prediction.}
\label{mattab: suppl_head}
\end{table}

\myparagraph{Our pose estimator}
Here we detail our pose estimator in\reffig{matfig: pose_est}. Given the fused features $\M{F'}$, the network predicts the 2D keypoints $\{(x_i, y_i)\}_{i=1}^{2J}$, the root-relative depth $z_i$ for each joint $i$ out of $2J$ joints, the handedness of the image $(h^L, h^R)\in \R^2$, and the relative depth between the left and the right roots $z^{R\rightarrow L}\in \R$. 
To obtain the 2D latent heatmaps $\M{H}^*_{2D}\in \R^{W_F\times H_F\times 2J}$ and latent depth maps $\M{H}^*_{z}\in \R^{W_F\times H_F\times 2J}$, which encode the 2D position and the root-relative depth for each joint, we perform $1\times 1$ convolution on $\M{F'}$. The 2D latent heatmaps $\M{H}^*_{2D}$ are softmax-normalized spatially into $\M{H}_{2D}$ such that each slice $\M{H}^i_{2D}$ for the joint $i$ sums to 1. Since $\M{H}_{2D}$ represent potential 2D joint locations, $\M{H}_{2D}$ is element-wise multiplied with the latent depth map $\M{H}^*_{z}$ to obtain the depth map $\M{H}_{z} = \M{H}^*_{z} \odot \M{H}_{2D}$ to focus the depth values on the joint locations.
Finally, the 2D keypoints are obtained by a soft-argmax operation on the 2D heatmap $\M{H}^i_{2D}$ for each joint $i$:
\begin{align}\label{mateq: softargmax2d}
    (x_i, y_i) = \sum_{m=0}^{W_F-1} \sum_{n=0}^{H_F-1} \M{H}_{2D}^i[m, n] (m, n)
\end{align}
where $(m, n)\in \R^2$ represents potential 2D position on the heatmap  for a joint $i$. The term $\M{H}_{2D}^i[m, n] \in \R$ is the probability of the 2D coordinate $(m,n)$ being the 2D position of the joint $i$. 
Note that, for brevity, we use $\M{H}^*_{2D}$ to denote the 2D latent heatmaps for all joints and  $\M{H}_{2D}^i$ to denote the normalized 2D heatmap for a joint $i$.
The root-relative depth $z_i$ for each joint $i$ is defined as
\begin{align}
    z_i = \sum_{m=0}^{W_F-1} \sum_{n=0}^{H_F-1} \M{H}^i_{z}[m, n].
\end{align}

To estimate the handedness $(h^L, h^R)\in \R^2$, and the relative depth between the two roots $z^{R\rightarrow L}\in \R$, we need to obtain a vector representation $\V{x}$. Therefore, we iteratively convolve and downsample $\M{F'}$
and we take a mean over the spatial dimension to obtain a latent vector $\V{x}$ (see \reftab{mattab: suppl_head}).
Finally, we learn two separately multi-layer perceptron (MLP) networks to map the latent vector $\V{x}$ to the handedness $(h^L, h^R)$ and a 1D heatmap $\V{p}\in \R^{D_z}$ that is softmax-normalized, representing the probability distribution over $D_z$ possible values for $z^{R\rightarrow L}$. The final relative depth $z^{R\rightarrow L}$ is obtained by
\begin{align}\label{mateq: z_relative}
    z^{R\rightarrow L} = \sum_{k=0}^{D_z-1} k\:\V{p}[k].
\end{align}
We use two linear layers in the MLPs with dimensions of 512 and ReLU\ccite{relu} activation.

\newpage

\newpage

\newpage

\end{document}